\documentclass[journal]{IEEEtran}

\usepackage{amsmath,epsfig}
\usepackage{multirow}
\usepackage{graphicx}
\usepackage{amssymb}
\usepackage[ruled]{algorithm2e}
\usepackage{colortbl}
\usepackage{tabularx}
\usepackage{lipsum}
\usepackage{booktabs}
\usepackage{siunitx}
\usepackage[square, comma, sort&compress, numbers]{natbib}
\usepackage[colorlinks]{hyperref}

\ifCLASSINFOpdf
\else
\fi

\begin{document}
%
% paper title

\title{FPGA: Fast Patch-Free Global Learning Framework for Fully End-to-End Hyperspectral Image Classification}

\author{{Zhuo~Zheng,~\IEEEmembership{Student Member,~IEEE,}
      Yanfei~Zhong,~\IEEEmembership{Senior Member,~IEEE,}
      Ailong~Ma,
      and~Liangpei~Zhang,~\IEEEmembership{Fellow,~IEEE,}}% <-this % stops a space

  \thanks{
    This work was supported by National Key Research and Development Program of China under Grant No. 2017YFB0504202, National Natural Science Foundation of China under Grant Nos. 41771385; and the National Natural Science Foundation of China under Grant NO. 41801267, in part by the China Postdoctoral Science Foundation under Grant 2017M622522.
    (\textit{Corresponding authors: Yanfei Zhong, Ailong Ma})

    The authors are with the State Key Laboratory of Information Engineering in Surveying, Mapping and Remote Sensing, Wuhan University, Wuhan 430079, China, with the Hubei Provincial Engineering Research Center of Natural Resources Remote Sensing Monitoring, Wuhan University, Wuhan 430079, China.
    (e-mail: zhengzhuo@whu.edu.cn; zhongyanfei@whu.edu.cn; maailong007@whu.edu.cn; zlp62@whu.edu.cn)
  }}

% The paper headers
\markboth{IEEE TRANSACTIONS ON GEOSCIENCE AND REMOTE SENSING}%
{Shell \MakeLowercase{\textit{et al.}}: Bare Demo of IEEEtran.cls for IEEE Journals}

% make the title area
\maketitle

% As a general rule, do not put math, special symbols or citations
% in the abstract or keywords.
\begin{abstract}
  % background
  Deep learning techniques have provided significant improvements in hyperspectral image (HSI) classification.
  The current deep learning based HSI classifiers follow a patch-based learning framework by dividing the image into overlapping patches.
  As such, these methods are local learning methods, which have a high computational cost.
  In this paper, a fast patch-free global learning (FPGA) framework is proposed for HSI classification.
  The proposed framework consists of three main parts: 1) a designed sampling strategy; 2) an encoder-decoder based fully convolutional network (FCN); and 3) lateral connections between the encoder and decoder.
  In FPGA, an encoder-decoder based FCN is utilized to consider the global spatial information by processing the whole image, which results in fast inference.
  However, it is difficult to directly utilize the encoder-decoder based FCN for HSI classification as it always fails to converge due to the insufficiently diverse gradients caused by the limited training samples.
  To solve the divergence problem and maintain the FCN’s abilities of fast inference and global spatial information mining, a global stochastic stratified (GS$^2$) sampling strategy is first proposed by transforming all the training samples into a stochastic sequence of stratified samples.
  This strategy can obtain diverse gradients to guarantee the convergence of the FCN in the FPGA framework. 
  For a better design of FCN architecture, FreeNet, which is a fully end-to-end network for HSI classification, is proposed to maximize the exploitation of the global spatial information and boost the performance via a spectral attention based encoder and a lightweight decoder. 
  A lateral connection module is also designed to connect the encoder and decoder, fusing the spatial details in the encoder and the semantic features in the decoder.
  The experimental results obtained using three public benchmark datasets suggest that the FPGA framework is superior to the patch-based framework in both speed and accuracy for HSI classification.
  Code has been made available at: \url{https://github.com/Z-Zheng/FreeNet}.
\end{abstract}

% Note that keywords are not normally used for peerreview papers.

% For peer review papers, you can put extra information on the cover
% page as needed:
% \ifCLASSOPTIONpeerreview
% \begin{center} \bfseries EDICS Category: 3-BBND \end{center}
% \fi
%
% For peerreview papers, this IEEEtran command inserts a page break and
% creates the second title. It will be ignored for other modes.
\IEEEpeerreviewmaketitle

\section{Introduction}
% The very first letter is a 2 line initial drop letter followed
% by the rest of the first word in caps.
% 
% form to use if the first word consists of a single letter:
% \IEEEPARstart{A}{demo} file is ....
% 
% form to use if you need the single drop letter followed by
% normal text (unknown if ever used by the IEEE):
% \IEEEPARstart{A}{}demo file is ....
% 
% Some journals put the first two words in caps:
% \IEEEPARstart{T}{his demo} file is ....
% 
% Here we have the typical use of a "T" for an initial drop letter
% and "HIS" in caps to complete the first word.
\label{sec:intro}
\IEEEPARstart{H}{yperspectral} imaging, as a particularly important technique, is able to obtain abundant spectral information about the ground surface \cite{ghamisi2018new,zhong2018mini, li2019deep}.
As a result, it is widely applied in the fields of geology, agriculture, forestry, and environmental monitoring \cite{govender2007review, adam2010multispectral,koch2010status}.
Hyperspectral image classification with the goal to assign a unique semantic label to each pixel in a hyperspectral image (HSI) \cite{camps2014advances}, is a fundamental but challenging part of hyperspectral remote sensing (HRS).

For HSI classification, the spectral feature-based methods, such as support vector machine (SVM) \cite{melgani2004classification}, random forest (RF) \cite{gislason2006random},  rotation forest (RoF) \cite{xia2013hyperspectral},  canonical correlation forest (CCF) \cite{rainforth2015canonical, xia2016hyperspectral} and multinomial logistic regression (MLR) \cite{krishnapuram2005sparse}, are the traditional classifiers for HSIs.
To further improve the accuracy of HSI classification, spatial information has been integrated into the existing pipelines \cite{fauvel2007spectral, tarabalka2009spectral, li2012spectral}.
Thereby, spectral-spatial feature based methods, such as the gray level cooccurrence matrix \cite{pesaresi2008robust}, wavelet transform \cite{zhu1998study}, the Gabor filter \cite{li2014gabor}, etc., have been proposed to improve the discrimination of the features.
Extended morphological profiles (EMPs) \cite{benediktsson2005classification, li2013generalized} has also been proposed to leverage the spatial context with multiple morphological operations for HSI classification.
However, these spectral-spatial features are handcrafted, and they are strongly reliant on the prior information and empirical hyperparameters \cite{chen2014deep, xu2018spectral}.

To automatically obtain more general spectral-spatial features, deep learning technology \cite{lecun2015deep}, as a data-driven automatic feature learning framework, has now been introduced into HSI classification.
Among the deep learning based methods, convolutional neural networks (CNNs), as hierarchical spectral-spatial feature representation learning frameworks, have been widely used in HSI classification \cite{yue2015spectral, hu2015deep, chen2016deep,yu2017convolutional, paoletti2018new}, significantly boosting the accuracy when compared with the traditional methods.
More importantly, CNN-based methods can act as an end-to-end training feature extractor and classifier for global optimization to obtain a better accuracy.
These CNN-based methods follow a patch-based local learning framework \cite{chen2014deep, yue2015spectral,paoletti2018new, xu2018spectral, zhu2018deformable, gong2019cnn,hang2019cascaded}, where patch generation is first performed to obtain a dense set of patches with a fixed size $S\times S$ and then patchwise classification is applied to each patch.

However, these methods usually have a high computational complexity under the patch-based local learning framework.
This is because these methods first generate overlapping image patches and then assign semantic labels obtained by the CNN to the corresponding central pixels to obtain complete classification map.
However this results in redundant computation since the image patches generated by adjacent pixels overlap with each other.
This seriously constrains the speed of the methods under the patch-based local learning framework.
Meanwhile, the limited patch size constrains the spatial context, making it difficult for the CNN to model long-range dependency.

In this work, a fast patch-free global learning (FPGA) framework is proposed for HSI classification.
The FPGA framework includes a sampling strategy, an encoder-decoder based FCN, and lateral connections between the encoder and decoder.
To share the computation in the spatial dimension and leverage the global spatial information, an encoder-decoder based fully convolutional network (FCN) is introduced to end-to-end HSI classification.
However, training an FCN for HSI classification is difficult since it always fails to converge.
The main reason for this is the insufficiently diverse gradients caused by the limited training samples during the backward computation.
To guarantee the convergence of the training of the FCN, a global stochastic stratified (GS$^2$) sampling strategy is proposed to obtain diverse gradients during back-propagation.
Furthermore, FreeNet, which is a novel network architecture for HSI classification, is proposed to maximize the exploitation of the global spatial information and further boost the performance.
FreeNet consists of a spectral attention based encoder and a lightweight decoder.
In addition, a lateral connection is applied to fuse the spatial details in the encoder and the semantic features in the decoder, for better exploitation of the encoder-decoder structural characteristics, which can help to recover more clear edges of objects in the classification map.

The main contributions of our study are summarized as follows:
\begin{enumerate}
  \item A fast patch-free global learning (FPGA) framework is proposed for HSI classification. The FPGA framework includes a sampling strategy, an encoder-decoder based FCN, and lateral connections between the encoder and decoder, which can achieve faster patch-free inference and learn from the global spatial information, for a better accuracy.
  \item To guarantee the convergence of the training of the FCN, the GS$^2$ sampling strategy is designed to assist with the training of the FCN. GS$^2$ strategy transforms all the training samples into a stochastic sequence of stratified samples, to obtain diverse gradients during back-propagation, for more effective parameter updating.
  \item To further boost the performance, a novel network architecture, FreeNet, is proposed for HSI classification through exploiting the global spatial information. FreeNet consists of a spectral attention based encoder and a lightweight decoder. Spectral attention involves modeling the interdependencies of the feature maps, using the global spatial context to guide the importance of the feature maps. This ensures sufficient exploitation of the redundant spectral information and the global spatial information. A lightweight decoder is responsible for the progressive recovery of the classification map, with less burden on optimization.
  \item To make full use of the encoder-decoder structural characteristics, a lateral connection between the encoder and decoder is designed for the fusion of the spatial details in the encoder with the semantic information in the decoder. This refines the semantic features with the spatial detail features to obtain a clearer classification map.
\end{enumerate}

The rest of this paper is organized as follows.
Section~\ref{sec:relate} briefly introduces the handcrafted feature based and CNN based HSI classifiers via the patch-based local learning framework.
Section~\ref{sec:upl} then describes the details of the proposed unified patch-free learning framework.
Section~\ref{sec:exp} describes the comparative results obtained on three HSI classification benchmark datasets, and further analyzes the proposed modules and the introduced hyperparameters.
Finally, Section~\ref{sec:conclusion} concludes this paper.

\section{Related Works}
\label{sec:relate}

\subsection{Deep Learning Based HSI Classifiers Under a Patch-Based Local Learning Framework}
\begin{figure}[h]
  \centering
  \includegraphics[width=\linewidth]{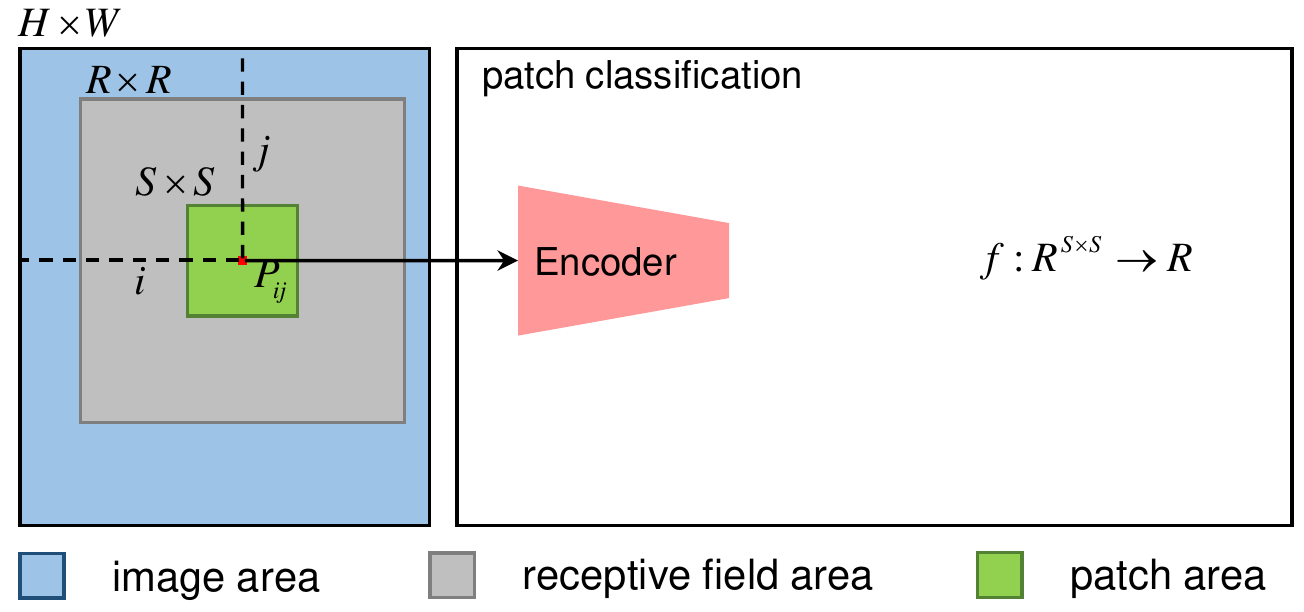}
  \caption{The patch-based local learning framework for HSI classification. For simplification, the mapping considers only the spatial dimension.}
  \label{fig:pbl}
\end{figure}

The dominant methods in modern HSI classification are based on deep networks and follow a patch-based local learning framework.
The deep networks include stacked autoencoders (SAEs), deep belief networks (DBNs), CNNs, recurrent neural networks (RNNs) and generative adversarial networks (GANs), which have all been explored in HSI classification \cite{chen2014deep, chen2015spectral,hu2015deep,chen2016deep, zhao2016spectral,xu2018spectral, gong2019cnn,hang2019cascaded, zhou2019learning}.
Among these methods, CNN-based classifiers, which are regarded as the natural spectral-spatial classification methods, have obvious advantages in accuracy.
To conveniently extract features and learn a classifier using CNNs, HSI patches are first generated from the original image by a window with a fixed size $S \times S$ (e.g. 7$\times$7 or 28$\times$28).
HSI classification always involves modeling a patch classification task \cite{ghamisi2018new, li2019deep}, which involves learning a mapping $f: R^{S\times S}\rightarrow R$, as shown in Fig.~\ref{fig:pbl}.

\begin{figure}
  \centering
  \includegraphics[width=0.7\linewidth]{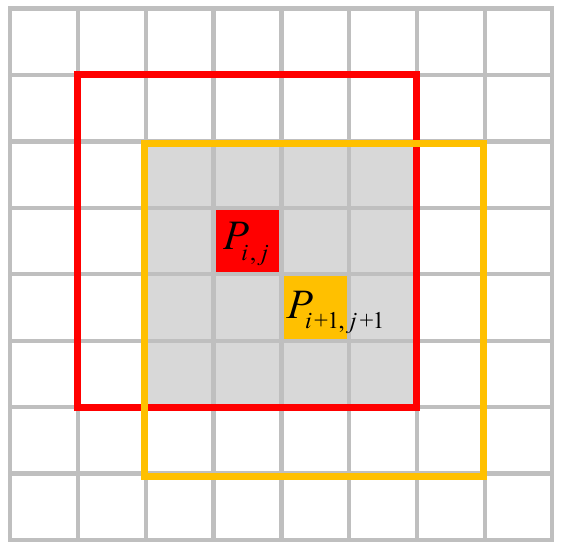}
  \caption{The overlap of 5$\times$5 patches under the patch-based local learning framework. The gray cell presents the overlap area. $P_{i,j}$ and $P_{i+1, j+1}$ are the central pixels of the two patches with red and yellow borders, respectively.}
  \label{fig:overlap}
\end{figure}

Under the patch-based local learning framework, the main difference between the CNN-based classifiers is in the different deep CNN designs.
A simple deep CNN was employed for HSI classification in \cite{hu2015deep}.
To utilize the spatial context information, a contextual CNN was proposed in \cite{lee2016contextual}, further improving the classification accuracy.
A CNN with pixel-pair features (CNN-PPF) was proposed in \cite{li2016hyperspectral} to enhance the original CNN by using deep pixel-pair features.
A spectral-spatial feature based classification framework was proposed in \cite{zhao2016spectral}, combining a balanced local discriminant embedding algorithm used as the spectral feature extractor with the CNN used as a spectral-spatial feature extractor for HSI classification.
To obtain more discriminative features, the Siamese convolutional neural network (S-CNN) was proposed in \cite{liu2017supervised} to learn low intraclass and high interclass features via a two-branch network, supervised by a margin ranking loss.
The deformable HSI classification networks (DHCNet) method \cite{zhu2018deformable} uses an adaptive spatial context modeling method to capture the complex spatial context in the HSIs and boost the performance.
However, the overfitting issue gradually emerges as the model complexity increases.
To alleviate this issue, Gabor-CNN \cite{chen2017hyperspectral} combines Gabor filters with convolutional filters to reduce the feature extraction burden of the CNN.
The deep feature fusion network (DFFN) was proposed in \cite{song2018hyperspectral} as a multi-layer feature fusion method that adopts residual learning to mitigate the overfitting brought by the introduction of more convolutional layers, significantly improving the classification accuracy for HSIs.
Although these patch-based methods have achieved remarkable HSI classification accuracies,
obtaining a fast inference speed remains a challenge, which limits the further application of HSI classifiers.
The main reason for this is the redundant computation in the overlapping areas between patches, as shown in Fig.~\ref{fig:overlap}.
The pixels in the gray area will take part in multiple computations since these pixels are the neighbors of multiple central pixels.

Although the DMS$^3$FE-classifier \cite{jiao2017deep} utilizes a pretrained FCN \cite{long2015fully} to extract features, its own FCN is not trained.
A convolution-deconvolution (conv-deconv) network with an optimized extreme learning machine (ELM) method was proposed in \cite{li2018classification} for HSI classification with an FCN, but the method is not an end-to-end classifier.

Neither of these methods are end-to-end trainable FCNs since they only utilize the FCN to extract features, and apply separate classifiers to label the pixels, which means that the whole pipeline cannot be globally optimized and the global spatial information cannot be exploited sufficiently.
To overcome the aforementioned issues, we propose the FPGA framework for HSI classification.

\section{FPGA: Fast Patch-Free Global Learning Framework for HSI Classification}
\label{sec:upl}
We investigated the main speed bottleneck of the patch-based methods and concluded that the redundant computation on the highly overlapping areas between patches is the central cause.
To address this issue, we propose a fast patch-free global learning (FPGA) framework and a variant of an FCN (FreeNet) as a fundamental classification model in the FPGA framework through sharing computation in the spatial dimension.

\begin{figure}[ht]
  \centering
  \includegraphics[width=\linewidth]{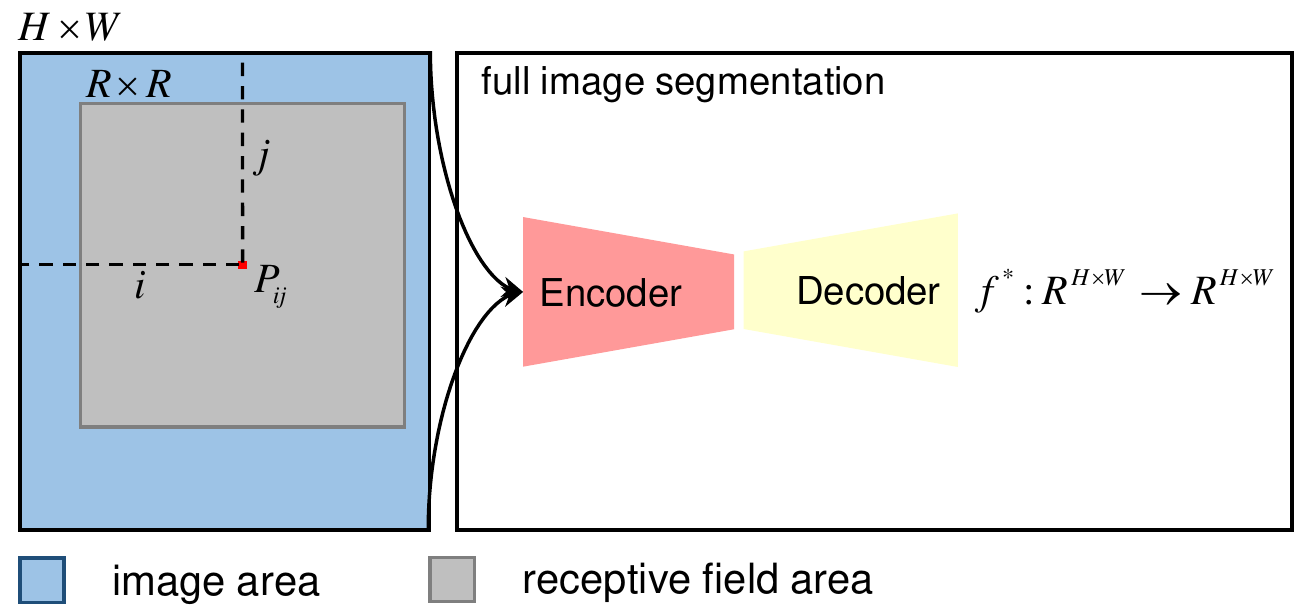}
  \caption{The patch-free global learning framework for HSI classification.
    For simplification, the mapping only considers the spatial dimension.}
  \label{fig:pfl}
\end{figure}

\begin{figure*}[ht]
  \centering
  \includegraphics[width=0.9\linewidth]{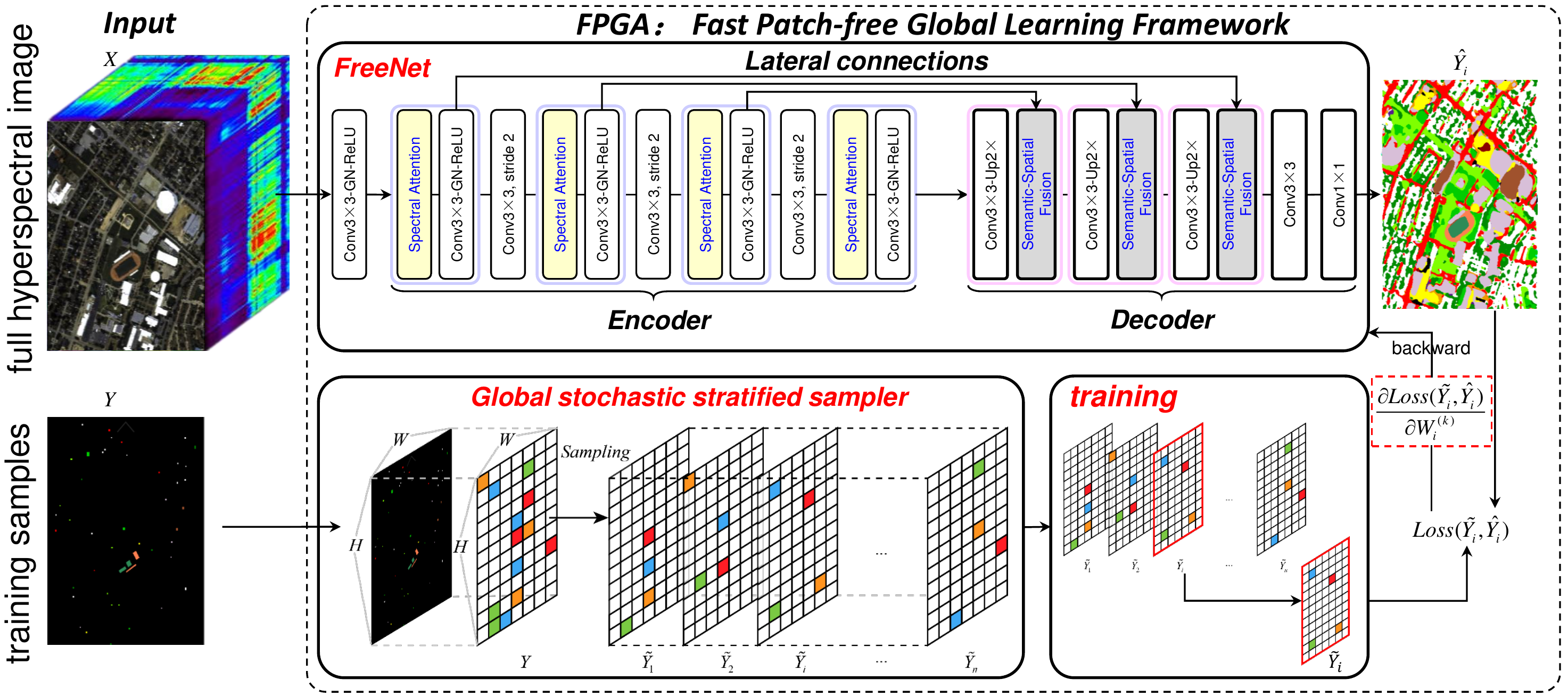}
  \caption{The overview of the fast patch-free global learning framework. The FPGA framework includes three core components: the GS$^2$ sampler, the encoder-decoder based FCN and lateral connections.}
  \label{fig:pipeline}
\end{figure*}

The FPGA framework aims to learn a mapping $f^*: R^{H\times W} \rightarrow R^{H\times W}$ for full image segmentation, as shown in Fig.~\ref{fig:pfl}.
In the FPGA framework, there are three core components: 1) the GS$^2$ sampler; 2) the encoder-decoder based FCN; and 3) the lateral connections.
The GS$^2$ sampler ensures the convergence of the end-to-end trained FCN based model, and the encoder-decoder based FCN is responsible for one-shot forward computation by sharing the computation in the spatial dimension.
The lateral connections are designed to effectively fuse the spatial detail features in the encoder and the semantic features in the decoder.
The model architecture follows the classical encoder-decoder framework \cite{long2015fully, ronneberger2015u, badrinarayanan2017segnet, chen2017deeplab} with semantic-spatial fusion (SSF).
The encoder is used to transform spatially finer features to semantically stronger features by progressively learning higher-dimensional feature embedding.
The decoder is used to recover the spatial information of the semantic features with the high-dimension feature embedding learned by the encoder for the full classification map.
The lateral connection based SSF forwards spatially finer features from the encoder to the decoder, which is beneficial for the recovery of the spatial details of the semantic feature maps.

\subsection{Patch-Free Global Learning}
The core idea of patch-free global learning is to replace the explicit patching with the implicit receptive field of the model, to avoid redundant computation on the overlapping areas and obtain a wider latent spatial context.

Given an HSI $X \in R^{C\times H\times W}$, the predicted probability cube $\hat{Y_i} \in R^{\#class\times H\times W}$ is formulated as:
\begin{equation}
  \hat{Y_i} = f^*(X)
\end{equation}
where the mapping $f^*: R^{C\times H\times W} \rightarrow R^{\#class\times H\times W}$ is modeled as the classifier without patching, and $C$ is the number of bands of $X$.

Stochastic gradient descent (SGD) \cite{robbins1951stochastic} is used to minimize the classification loss $l$ (e.g., cross-entropy loss) over the sampled positions $\mathcal{R}$.
The sampling algorithm is described in Section.~\ref{sec:gs2_sampler}.

For the $i$-th iteration, the $k$-th weight of the model can be updated as follows:
\begin{equation}
  W^{(k)}_{i+1} = W^{(k)}_{i} - \eta \frac{1}{n}\sum_{p \in \mathcal{R}_i}{\frac{\partial{l(\widetilde{Y}_i(p), \hat{Y_i}(p))}}{\partial{W^{(k)}_{i}}}}
\end{equation}
where $p$ is the 2-D spatial position in $\mathcal{R}_i$, $n = |\mathcal{R}_i|$, $\eta$ is the learning rate and $\widetilde{Y}_i$ is the sampled ground truth map.

The main difference with patch-based local learning is that all the pixels take part in the forward computation during the training, but only the sampled position can obtain supervised signals for every iteration.
In this way, the model inference is consistent during the training and testing, which are both one-shot forward computations.
The one-shot forward computation significantly boosts the speed of the model inference for HSI classification.
Meanwhile, the patch-free global learning allows the model to leverage the spatial context as much as possible.
It thus provides more potential to boost the accuracy of the model by following the patch-free global learning framework.

\subsection{Global Stochastic Stratified Sampling Strategy}
\label{sec:gs2_sampler}
The global stochastic stratified (GS$^2$) sampling strategy is proposed to ensure the convergence of the end-to-end trained FCN based model.
The GS$^2$ sampling strategy is formally described in Algorithm \ref{alg:sss}.
The key idea of the GS$^2$ sampler is to transform all the training samples into a stochastic sequence of stratified samples.
In this way, the GS$^2$ sampler ensures the class-balanced distribution of the training samples and simulates the behavior of the mini-batch sampler to obtain stable yet diverse gradients.
During the training, this sequence is randomly shuffled to ensure stochasticity of the gradients, to prevent overfitting.

\begin{algorithm}[t]
  \caption{Global Stochastic Stratified Sampling\label{alg:sss}}
  \KwIn{$G = \{g_i\}_{i=1}^{M}$: a set of labels for training  \\
  \hspace{0.38in} \boldmath$N$: the number of classes \\
  \hspace{0.38in} \boldmath$\alpha$: mini-batch per class }
  \KwOut{$T$: a list of sets of stratified labels}
  $R \gets [] $ // an empty list \\
  \For{$k=0$ \KwTo $N$}{
    $I_k$ $\gets$ \{$j | g_j = k, g_j \in G$\} \\
    $I_k \gets$ shuffle($I_k$) \\

    $R[k] \gets []$ \\
    \While{$|I_k| > \alpha$}{
      fetch $\alpha$ samples from $I_k$, $t \gets I_k.pop(\alpha)$ \\
      $R[k].push(t)$
    }
    $R[k].push(I_k)$ \\
  }
  $T \gets []$ \\
  $c \gets 0$ \\
  \While{$any(R[i] > 0, i = 1,2,...,N-1)$}{
  $T[c] \gets \emptyset$ \\
  \For{$k=0$ \KwTo $N$}{
  \If{$|R[k]| > 0$}{
  fetch 1 element from $R[k]$, $t_k = R[k].pop(1)$ \\
  $T[c] \gets T[c]\cap t_k$ \\
  }

  }
  $T.push(T[c])$ \\
  $c \gets c+1 $
  }
\end{algorithm}

Firstly, in more detail, we split all the training samples to obtain a list $R$, where the index of $R$ is the class label, and the element is the training samples of each class.
During the splitting, the order of the training samples for each class needs to be shuffled to keep the stochasticity of the combination.
Stratification is then performed on the training samples of each class to obtain $T$, which is a list of the sets of stratified labels.

In this sampling strategy, the hyperparameter $\alpha$ (the mini-batch per class) is introduced, which is of great significance for the training stability.
The smaller the value of $\alpha$, the greater the number of gradient orientations that can be obtained when optimizing the network, which makes it possible to train an FCN using limited training samples for HSI classification.
A more detailed analysis of parameter $\alpha$ is provided in Section \ref{sec:gs2}.

\subsection{FreeNet in FPGA}
\begin{figure}[hbt]
  \centering
  \includegraphics[width=0.8\linewidth]{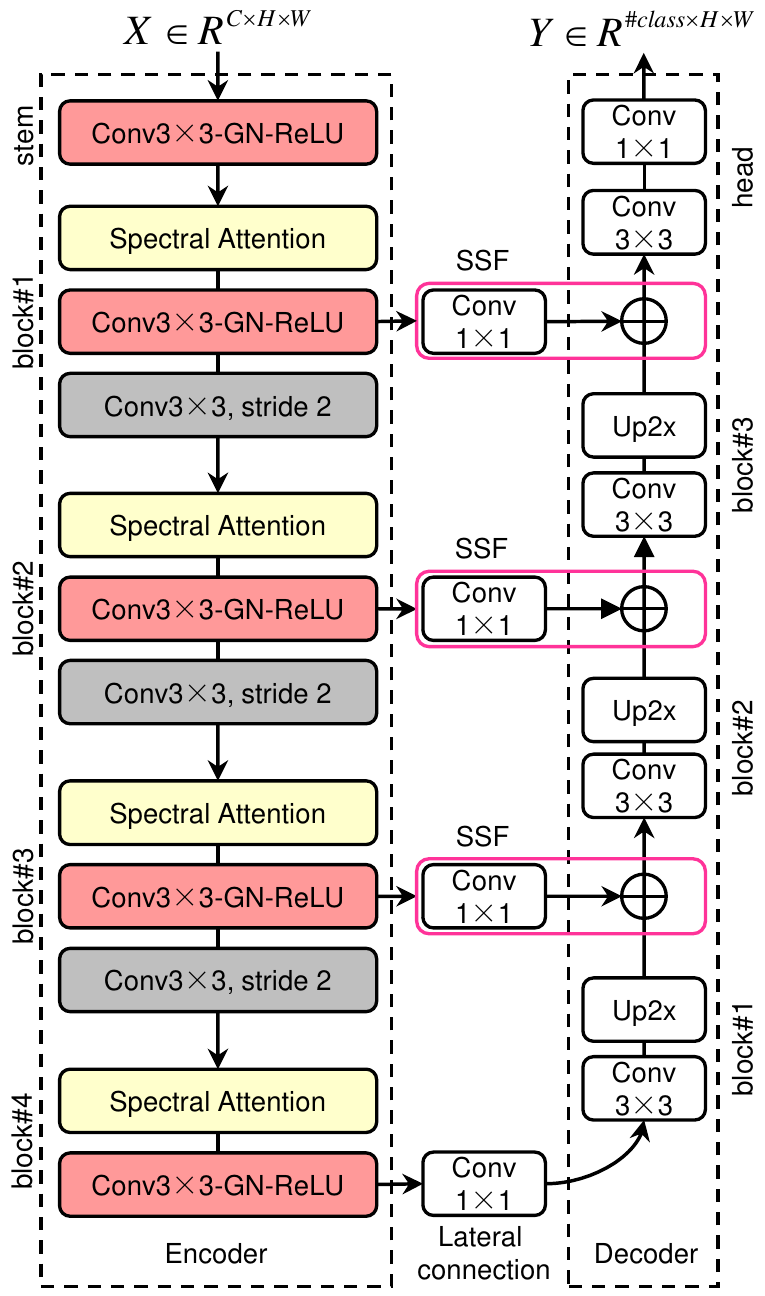}
  \caption{The FreeNet network architecture designed for HSI classification.}
  \label{fig:freenet}
\end{figure}

FreeNet is a simple, unified network made up of an \textit{encoder} network and a \textit{decoder} network.
The encoder is responsible for computing the hierarchical convolutional feature maps over an entire input HSI.
The decoder recovers the spatial dimension of the coarsest convolutional feature map progressively via lateral connection based SSF, outputting a classification probability map of the same spatial size as the input image.
To improve the FreeNet compactness, we introduce a compression factor ($\beta$) to control the number of feature maps in the whole network, achieving a trade-off between speed and accuracy.
FreeNet is a lightweight FCN designed for faster and more accurate HSI classification, as shown in Fig.~\ref{fig:freenet}.
Each component of FreeNet is described in the following.
\subsubsection{\textbf{Encoder Network Architecture}}
The encoder network follows a modular design, made up of a stem block and four hybrid blocks, all of which contain the basic module.
The basic module of the encoder network is a 3$\times$3 convolutional layer followed by group normalization \cite{wu2018group} and rectified linear unit (ReLU) activation.
Under the FPGA framework, the batch size is always equal to 1, and the iterative inputs are the same image, because of the entire HSI being used as input.
In this case, the error of the batch normalization (BN) increases rapidly due to the inaccurate batch statistics estimation.
Therefore, we adopt group normalization (GN) as an alternative to BN, which is independent of batch size and can obtain a comparable performance to BN.

Due to the different numbers of bands for HSIs, we first introduce a stem block to transform the variable channels of the input to a fixed 64 channels.
The stem block is simply implemented by a basic module.
The four hybrid blocks share the same network topology, unless otherwise specified.
The hybrid block is made up of a spectral attention module, as described in Section.~\ref{sec:sa}, a basic module and an optional downsampling module.
For the downsampling module, we use a 3$\times$3 convolutional layer with a stride of 2 followed by ReLU activation to replace the commonly used 3$\times$3 maxpool with a stride of 2 to align the projected spatial location with its receptive field center for more robust HSI classification.
Block\#1 $\sim$ \#3 are the hybrid blocks with downsampling modules and block\#4 is the block without a downsampling module.

\subsubsection{\textbf{Spectral Attention}}
\label{sec:sa}
Spectral attention models the interdependencies of the feature maps with the global spatial context.
The encode function is simply implemented by $3\times 3$ convolution.
This module reweights the feature maps via global context guiding and highlights the more important feature maps to improve the accuracy of the model.
\begin{figure}[hbt]
  \centering
  \includegraphics[width=\linewidth]{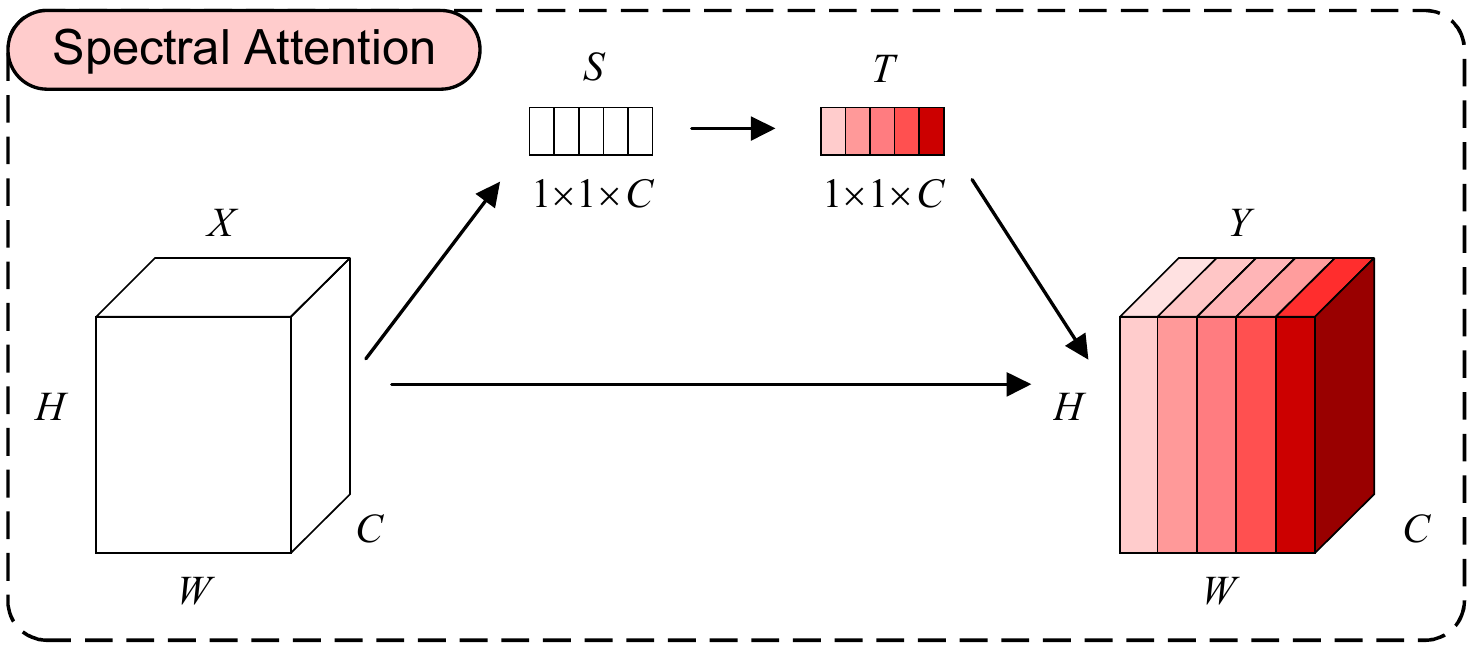}
  \caption{The network architecture of the spectral attention module.}
  \label{fig:sa}
\end{figure}

The spectral attention is an in-place module where there is no dimension change between the input and output, which is a similar implementation to the SE-Block \cite{hu2018squeeze}.
Given the input tensor $X \in \mathbb{R}^{C\times H \times W}$, we first compute the global context embedding vector $S \in \mathbb{R}^{C\times 1 \times 1}$, where
\begin{equation}
  S(k,:, :) =  \frac{1}{H \cdot W} \sum_{i=1}^{H}\sum_{j=1}^{W}X(k, i, j)
\end{equation}
where $k$ is the index of the channel dimension.
In order to model the interdependencies between feature maps, we use two thin fully connected layers (nonlinear transformation), followed by a sigmoid gating function to compute the channel scaling coefficient vector $T \in \mathbb{R}^{C\times 1 \times 1}$, where
\begin{equation}
  T = sigmoid(W_2\delta(W_1S))
\end{equation}
where $\delta$ denotes the ReLU function, $W_1 \in \mathbb{R}^{\frac{C}{r} \times C}$, and $W_2 \in \mathbb{R}^{C \times \frac{C}{r}}$.
The reduction ratio $r$ is introduced to balance the capacity of the model and the computational cost.
We find $r = 16$ works well in practice, so this setting was used in most of the experiments.
The output $Y \in \mathbb{R}^{C\times H \times W}$ of the spectral attention module is simply computed by:
\begin{equation}
  Y(k, i, j) = T(k,:, :) \cdot X(k, i, j)
\end{equation}

\subsubsection{\textbf{Decoder Network Architecture}}
The decoder network also follows a modular design, for simplicity, which is made up of a refinement module for progressive spatial feature refinement and a head subnetwork for pixel classification.
The refinement module contains multiple refinement stages, which are simply implemented by stacking the upsampling modules and inserting SSF after each upsampling module.
The progressive refinement first upsamples the input feature maps with stronger semantic information and then aggregates the feature map with finer spatial information from the encoder to recover the spatial details of the input.
The upsampling module is a 3$\times$3 convolutional layer followed by nearest neighbor upsampling with a factor of 2.
The SSF receives two features from the blocks in the encoder and decoder, respectively, and aggregates the features into a new enhanced feature that is forwarded to the next upsampling module.
The head subnetwork is used to perform pixel classification with the feature from the top layer of the decoder, and is made up of a 3$\times$3 convolutional layer followed by a 1$\times$1 convolutional layer with $N$ filters.
$N$ is the number of categories.

\subsubsection{\textbf{Lateral Connection Based SSF}}
\label{sec:ssf}
Lateral connection based SSF leverages the spatial detail features of the shallow convolutional layers to enhance the deep semantic features of the convolutional layers and boost the performance.
The lateral connection is implemented by a 1$\times$1 convolutional layer, which passes more precise locations of features from the encoder to the decoder.
The aggregation function is pointwise addition.
Lateral connection based SSF can be formulated as follows:
\begin{equation}
  q_{i+1} = q_i + conv(p_{4-i}), i = 1,2,3
\end{equation}
where $q_i$ is the feature map of refinement stage\#$i$ in the decoder, and $p_{4-i}$ is the feature map of block\#$4-i$ in the encoder.
$q_{i+1}$ is the output of SSF, which is forwarded to the next block in the decoder.
The design of the lateral connection based SSF follows that of residual learning \cite{he2016deep}.
The first item $q_i$ is a baseline item obtained by nearest neighbor interpolation and $conv(p_{4-i})$ is the residual item that needs to be learned.
In this case, the gradients are lossless to flow into the shallow layers, making the optimization easier.

\subsection{Fully End-to-End HSI Classification Using FreeNet in FPGA}
After convergence of the trained FreeNet, HSI classification can be implemented by one-shot forward computation using FreeNet in FPGA.
Compared with patchwise classification for HSIs, FreeNet can perform faster patch-free inference over the whole HSI through sharing the computation in the spatial dimension.
Due to the introduction of three $2\times$ upsampling blocks in FreeNet, the input size in the spatial dimension of the HSI should be multiples of $2^3 = 8$.
To ensure that the raw HSI is unchanged, a ``padding-crop'' trick is used for the inference, which pads the original input with zeros into a new size with multiples of 8 before the inference.
After the inference, the final classification map is obtained by cropping the output using a box with the original input size.

\section{Experimental results and analysis}
\label{sec:exp}
Extensive experiments were conducted to validate the effectiveness of the proposed method on three benchmark datasets: the ROSIS-03 Pavia University dataset, the Salinas dataset and the Compact Airborne Spectrographic Imager (CASI) University of Houston dataset.
Four patch-based HSI classifiers were compared with the proposed FreeNet and its variants.
The patch-based classifiers used in the comparison were classical SVM \cite{melgani2004classification} and four state-of-the-art deep learning based methods (S-CNN \cite{liu2017supervised}, Gabor-CNN \cite{chen2017hyperspectral}, DFFN \cite{song2018hyperspectral}, 3D-GAN \cite{zhu2018generative}), following \cite{li2019deep}.
All the experiments were performed with an NVIDIA Tesla P100 GPU accelerator (with 16GB GPU memory).

\begin{table}[htb]
  \caption{The configuration details of the standard FreeNet (FreeNet with $\beta=1.0$).
    \label{tab:freenet_setting}}
  \centering
  \renewcommand{\arraystretch}{1.5}
  \begin{tabular}{ccc}
    \hline
    \multicolumn{3}{c}{FreeNet ($\beta=1.0$)}                                                                                        \\ \hline
    \multicolumn{1}{c|}{\multirow{12}{*}{Encoder}} & \multicolumn{1}{c|}{stem}                      & 3$\times$3 conv, 64            \\ \cline{2-3}
    \multicolumn{1}{c|}{}                          & \multicolumn{1}{c|}{\multirow{3}{*}{block\#1}} & spectral attention, $r = 16$   \\
    \multicolumn{1}{c|}{}                          & \multicolumn{1}{c|}{}                          & 3$\times$3 conv, 64            \\
    \multicolumn{1}{c|}{}                          & \multicolumn{1}{c|}{}                          & 3$\times$3 conv, 128, stride 2 \\ \cline{2-3}
    \multicolumn{1}{c|}{}                          & \multicolumn{1}{c|}{\multirow{3}{*}{block\#2}} & spectral attention, $r = 16$   \\
    \multicolumn{1}{c|}{}                          & \multicolumn{1}{c|}{}                          & 3$\times$3 conv, 128           \\
    \multicolumn{1}{c|}{}                          & \multicolumn{1}{c|}{}                          & 3$\times$3 conv, 192, stride 2 \\ \cline{2-3}
    \multicolumn{1}{c|}{}                          & \multicolumn{1}{c|}{\multirow{3}{*}{block\#3}} & spectral attention, $r = 16$   \\
    \multicolumn{1}{c|}{}                          & \multicolumn{1}{c|}{}                          & 3$\times$3 conv, 192           \\
    \multicolumn{1}{c|}{}                          & \multicolumn{1}{c|}{}                          & 3$\times$3 conv, 256, stride 2 \\ \cline{2-3}
    \multicolumn{1}{c|}{}                          & \multicolumn{1}{c|}{\multirow{2}{*}{block\#4}} & spectral attention, $r = 16$   \\
    \multicolumn{1}{c|}{}                          & \multicolumn{1}{c|}{}                          & 3$\times$3 conv, 256           \\ \hline
    \multicolumn{1}{c|}{\multirow{4}{*}{Lateral}}  & \multicolumn{1}{c|}{lateral 4-1}               & 1$\times$1 conv, 128           \\ \cline{2-3}
    \multicolumn{1}{c|}{}                          & \multicolumn{1}{c|}{lateral 3-1}               & 1$\times$1 conv, 128           \\ \cline{2-3}
    \multicolumn{1}{c|}{}                          & \multicolumn{1}{c|}{lateral 2-2}               & 1$\times$1 conv, 128           \\ \cline{2-3}
    \multicolumn{1}{c|}{}                          & \multicolumn{1}{c|}{lateral 1-3}               & 1$\times$1 conv, 128           \\ \hline
    \multicolumn{1}{c|}{\multirow{8}{*}{Decoder}}  & \multicolumn{1}{c|}{\multirow{2}{*}{block\#1}} & 3$\times$3 conv, 128           \\
    \multicolumn{1}{c|}{}                          & \multicolumn{1}{c|}{}                          & upsample, 2                    \\ \cline{2-3}
    \multicolumn{1}{c|}{}                          & \multicolumn{1}{c|}{\multirow{2}{*}{block\#2}} & 3$\times$3 conv, 128           \\
    \multicolumn{1}{c|}{}                          & \multicolumn{1}{c|}{}                          & upsample, 2                    \\ \cline{2-3}
    \multicolumn{1}{c|}{}                          & \multicolumn{1}{c|}{\multirow{2}{*}{block\#3}} & 3$\times$3 conv, 128           \\
    \multicolumn{1}{c|}{}                          & \multicolumn{1}{c|}{}                          & upsample, 2                    \\ \cline{2-3}
    \multicolumn{1}{c|}{}                          & \multicolumn{1}{c|}{\multirow{2}{*}{head}}     & 3$\times$3 conv, 128           \\
    \multicolumn{1}{c|}{}                          & \multicolumn{1}{c|}{}                          & 1$\times$1 conv, $N$           \\ \hline
  \end{tabular}
\end{table}

\begin{figure*}[htb]
  \begin{minipage}[b]{0.15\linewidth}
    \centering
    \includegraphics[width=\linewidth]{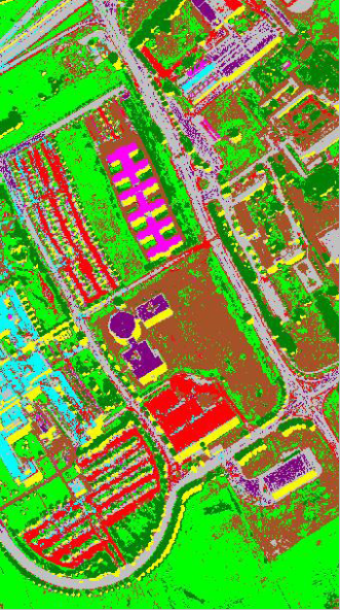}
    \centerline{(a) SVM}
  \end{minipage}
  \hfill
  \hspace{4pt}
  \begin{minipage}[b]{0.15\linewidth}
    \centering
    \includegraphics[width=\linewidth]{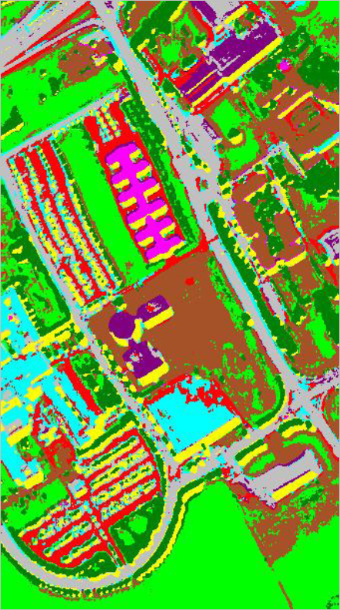}
    \centerline{(b) S-CNN}
  \end{minipage}
  \hfill
  \hspace{4pt}
  \begin{minipage}[b]{0.15\linewidth}
    \centering
    \includegraphics[width=\linewidth]{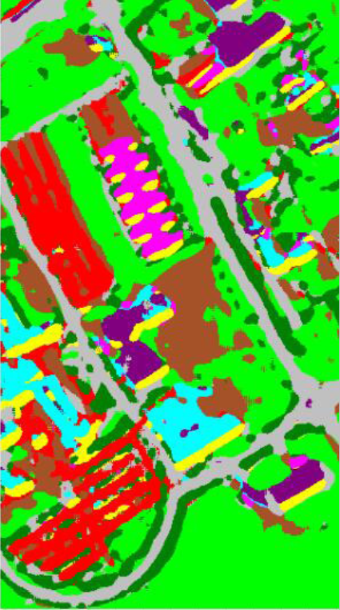}
    \centerline{(c) Gabor-CNN}
  \end{minipage}
  \hfill
  \hspace{4pt}
  \begin{minipage}[b]{0.15\linewidth}
    \centering
    \includegraphics[width=\linewidth]{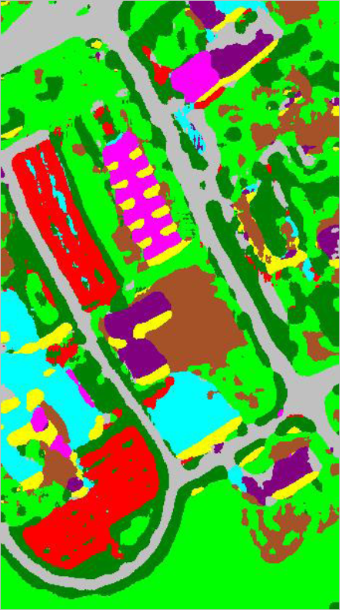}
    \centerline{(d) DFFN}
  \end{minipage}
  \hfill
  \hspace{4pt}
  \begin{minipage}[b]{0.15\linewidth}
    \centering
    \includegraphics[width=\linewidth]{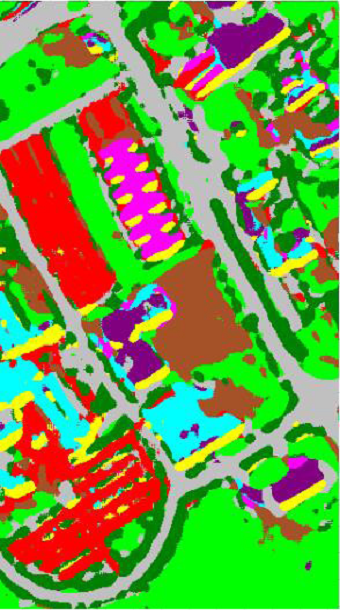}
    \centerline{(e) 3D-GAN}
  \end{minipage}
  \hfill
  \hspace{4pt}
  \begin{minipage}[b]{0.15\linewidth}
    \centering
    \includegraphics[width=\linewidth]{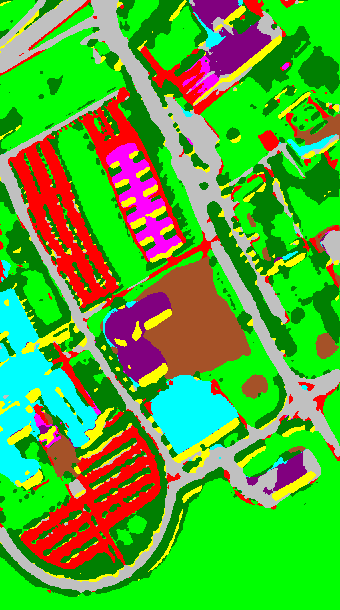}
    \centerline{(f) FreeNet}
  \end{minipage}
  \caption{Visualization of the classification maps for the ROSIS-03 Pavia University dataset.
    (a) SVM. (b) S-CNN. (c) Gabor-CNN. (d) DFFN. (e) 3D-GAN. (f) FreeNet.}
  \label{fig:vis_pavia}
\end{figure*}

\subsection{Experimental Settings}
\subsubsection{\textbf{Network Architecture}}
The hyperparameter settings of the standard FreeNet, namely FreeNet ($\beta=1.0$), such as the output channels of the layers and the reduction ratio $r$ in the spectral attention module, are listed in Table.~\ref{tab:freenet_setting}.
For a fair comparison, this architecture setting was used for all three benchmark datasets, and there was no specific tuning for the dataset.

\subsubsection{\textbf{Optimization}}
For all the experiments, the patch-free methods were trained for 1k iterations using SGD with a ``poly'' learning rate policy, where the initial learning rate was set to 0.0001 and multiplied by $(1 - \frac{iter}{max\_iter})^{power}$ with $power = 0.9$.
The momentum was set to 0.9 and the weight decay was set to 0.0001.
We did not use any data augmentation strategy.
Unless otherwise specified, $\alpha$ was set to 20 for the GS$^2$ sampler.

\subsubsection{\textbf{Metrics}}
To evaluate the performance of the proposed methods, four common metrics are adopted, which are the accuracy of each class, the overall accuracy (OA),  the average
accuracy (AA), and the Kappa coefficient (Kappa).

\subsection{Experiment 1: ROSIS-03 Pavia University Dataset}
\label{sec:exp1}

\begin{figure}[htb]
  \begin{minipage}[b]{.30\linewidth}
    \centering
    \includegraphics[width=\linewidth]{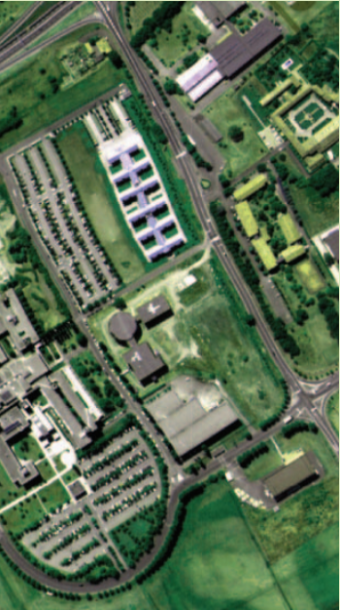}
    \centerline{(a)}
  \end{minipage}
  \hfill
  \begin{minipage}[b]{0.30\linewidth}
    \centering
    \includegraphics[width=\linewidth]{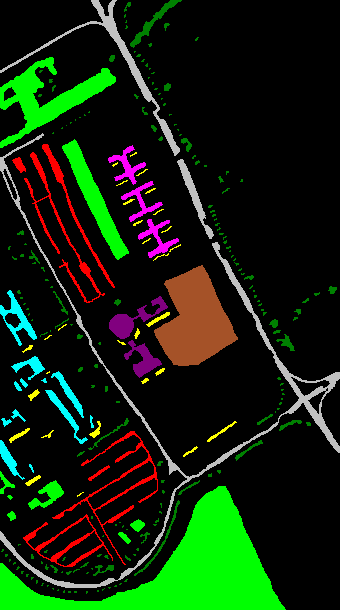}
    \centerline{(b)}
  \end{minipage}
  \hfill
  \begin{minipage}[b]{0.30\linewidth}
    \centering
    \includegraphics[width=\linewidth]{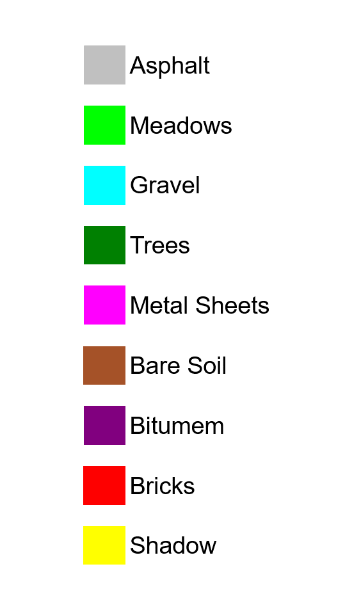}
    \centerline{(c)}
  \end{minipage}
  \caption{The ROSIS-03 Pavia University dataset.
    (a) Three-band false color composite.
    (b) Ground-truth map
    (c) Legend}
  \label{fig:pavia_data}
\end{figure}

The HSI of this dataset contains 610 $\times$ 340 pixels and 103 spectral bands with a spatial resolution of 1.3 m per pixel.
This dataset contains nine urban land-cover types.
Fig.~\ref{fig:pavia_data} shows the false-color composite of the image and the corresponding ground truth.

\begin{table}[htb]
  \caption{The number of training samples and test samples for the ROSIS-03 Pavia University dataset
    \label{tab:pavia_sample}}
  \centering
  \renewcommand{\arraystretch}{1.5}
  \begin{tabular}{l|l|c|c|c}
    \hline
    Class & Class name   & \#Training & \#Test & \#Total \\
    \hline
    C1    & Asphalt      & 200        & 6431   & 6631    \\
    C2    & Meadows      & 200        & 18449  & 18649   \\
    C3    & Gravel       & 200        & 1899   & 2099    \\
    C4    & Trees        & 200        & 2864   & 3064    \\
    C5    & Metal Sheets & 200        & 1145   & 1345    \\
    C6    & Bare Soil    & 200        & 4829   & 5029    \\
    C7    & Bitumem      & 200        & 1130   & 1330    \\
    C8    & Bricks       & 200        & 3482   & 3682    \\
    C9    & Shadow       & 200        & 747    & 947     \\
    \hline
    Total & -            & 1800       & 40976  & 42776   \\
    \hline
  \end{tabular}
\end{table}

\begin{table}[hbt]
  \caption{The classification results of SVM \cite{melgani2004classification}, S-CNN\cite{liu2017supervised}, Gabor-CNN \cite{chen2017hyperspectral}, DFFN \cite{song2018hyperspectral}, 3D-GAN \cite{zhu2018generative} and FreeNet on the ROSIS-03 Pavia University dataset.
    \label{tab:result_pavia}}
  \centering
  \renewcommand{\arraystretch}{1.5}
  \resizebox{\linewidth}{!}{
    \begin{tabular}{c|c|ccccc|c}
      \hline
      \multicolumn{2}{c|}{\multirow{2}{*}{Class}}   & \multicolumn{5}{c|}{Patch-based} & \multicolumn{1}{c}{Patch-free}                                                            \\ \cline{3-8}
      \multicolumn{2}{c|}{}                         & SVM                              & S-CNN                          & Gabor-CNN & DFFN  & 3D-GAN  & \multicolumn{1}{c}{FreeNet}         \\ \hline
      \multirow{9}{*}{\rotatebox{90}{Accuracy(\%)}} & C1                               & 85.49                          & 95.47     & 99.53         & 99.53 & 99.18                     & 99.58 \\
                                                    & C2                               & 92.12                          & 98.71     & 98.21  & 97.71   &   98.86                 & 99.88 \\
                                                    & C3                               & 85.77                          & 97.32     & 89.74  & 99.89   &  94.94                  & 99.95 \\
                                                    & C4                               & 96.41                          & 97.72     & 93.02  & 97.88   & 90.15                   & 99.27 \\
                                                    & C5                               & 98.60                          & 100       & 99.42  & 99.48   &  99.49                  & 100   \\
                                                    & C6                               & 92.52                          & 97.67     & 98.77  & 99.69   & 98.56                   & 100   \\
                                                    & C7                               & 93.79                          & 98.36     & 98.82  & 100     & 92.74                   & 100   \\
                                                    & C8                               & 86.56                          & 95.56     & 94.12  & 98.59  & 97.18                    & 99.83 \\
                                                    & C9                               & 97.97                          & 100       & 97.91  & 99.61   & 98.51                   & 100   \\ \hline
      \multicolumn{2}{c|}{OA(\%)}                   & 90.78                            & 97.93                          & 97.33     & 98.57 & 97.81 & 99.81                               \\
      \multicolumn{2}{c|}{AA(\%)}                   & 92.14                            & 97.88                          & 96.62     & 99.16 & 96.65 & 99.83                               \\
      \multicolumn{2}{c|}{Kappa}                    & 0.8813                           & 0.9743                         & 0.9662    & 0.9808&  0.9697 & 0.9974                              \\ \hline
    \end{tabular}
  }
\end{table}

Table.~\ref{tab:pavia_sample} lists the number of training and test samples for each class.
The training samples were randomly chosen from the ground truth with a fixed random seed and the remaining samples were used to evaluate the accuracy.

For the HSI classification task, the visual performance is of importance for the classifier.
Fig.~\ref{fig:vis_pavia} shows the classification maps of the compared methods.
As can be seen in Fig.~\ref{fig:vis_pavia} (b)-(f), the CNN based classifiers have a better visual performance than the classical SVM classifier due to the strongly discriminative deep features.
It is clear that the patch-free method is superior to the patch-based methods in spatial detail when comparing Fig.~\ref{fig:vis_pavia} (f) and Fig.~\ref{fig:vis_pavia} (b)-(e).
This can be attributed to the introduction of more global spatial context.
Among the different methods, the edges of the classification map of FreeNet are smoother than those of the other methods.
This indicates that FreeNet can capture finer spatial detail by the lateral connection based SSF and the better design of decoder structure.

Table.~\ref{tab:result_pavia} lists the results of the state-of-the-art patch-based methods and the proposed patch-free methods.
When constraining the number of training samples (200 samples for each class), DFFN achieves the best accuracy (OA of 98.57\%, AA of 99.16\%, and Kappa of 0.9808) among the patch-based methods.
However, the patch-free method obtains more accurate results than DFFN.
FreeNet obtains a higher OA of 99.81\%, exceeding DFFN by $\sim$1\%.
Meanwhile, we can observe that the accuracies for each class with the patch-free method are higher than those of the patch-based methods.
In fact, the accuracy of the patch-free method has reached saturation.
This suggests that the patch-free global learning framework is superior to the patch-based local learning framework with this benchmark dataset.

% ----------------------------------------------------------
\begin{figure}[hbt]
  \begin{minipage}[b]{.30\linewidth}
    \centering
    \includegraphics[width=\linewidth]{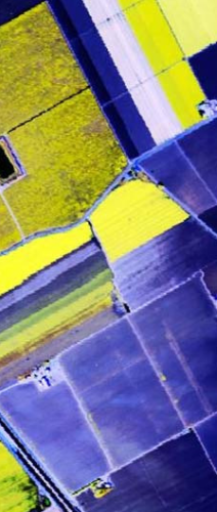}
    \centerline{(a)}
  \end{minipage}
  \hfill
  \begin{minipage}[b]{0.30\linewidth}
    \centering
    \includegraphics[width=\linewidth]{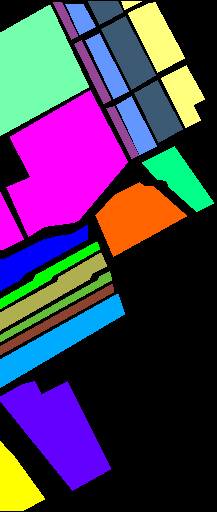}
    \centerline{(b)}
  \end{minipage}
  \hfill
  \begin{minipage}[b]{0.25\linewidth}
    \centering
    \includegraphics[width=\linewidth]{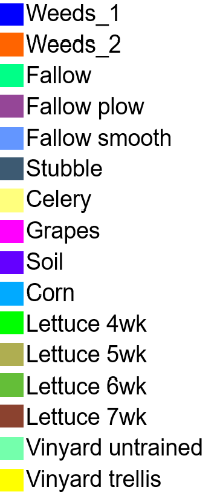}
    \centerline{(c)}
  \end{minipage}
  \caption{The Salinas dataset.
    (a) Three-band false color composite.
    (b) Ground truth map
    (c) Legend}
  \label{fig:salinas}
\end{figure}
\subsection{Experiment 2: Salinas Dataset}
\begin{figure*}[htb]
  \begin{minipage}[b]{0.15\linewidth}
    \centering
    \includegraphics[width=\linewidth]{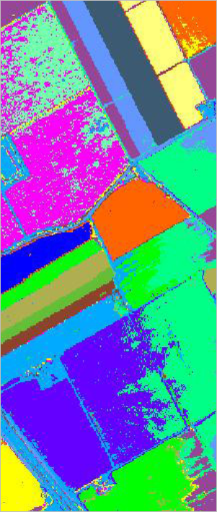}
    \centerline{(a) SVM}
  \end{minipage}
  \hfill
  \hspace{4pt}
  \begin{minipage}[b]{0.15\linewidth}
    \centering
    \includegraphics[width=\linewidth]{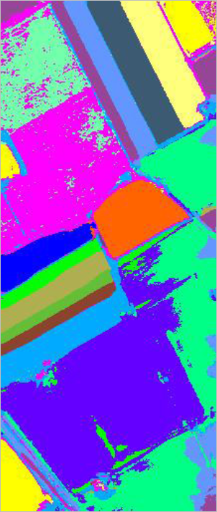}
    \centerline{(b) S-CNN}
  \end{minipage}
  \hfill
  \hspace{4pt}
  \begin{minipage}[b]{0.15\linewidth}
    \centering
    \includegraphics[width=\linewidth]{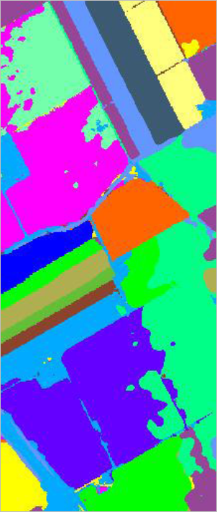}
    \centerline{(c) Gabor-CNN}
  \end{minipage}
  \hfill
  \hspace{4pt}
  \begin{minipage}[b]{0.15\linewidth}
    \centering
    \includegraphics[width=\linewidth]{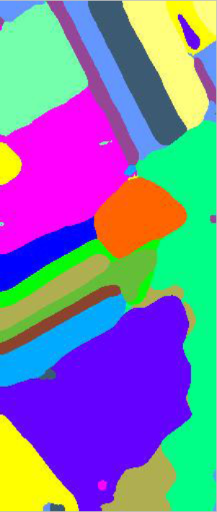}
    \centerline{(d) DFFN}
  \end{minipage}
  \hfill
  \hspace{4pt}
  \begin{minipage}[b]{0.15\linewidth}
    \centering
    \includegraphics[width=\linewidth]{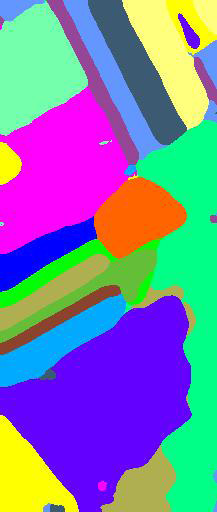}
    \centerline{(e) 3D-GAN}
  \end{minipage}
  \hfill
  \hspace{4pt}
  \begin{minipage}[b]{0.15\linewidth}
    \centering
    \includegraphics[width=\linewidth]{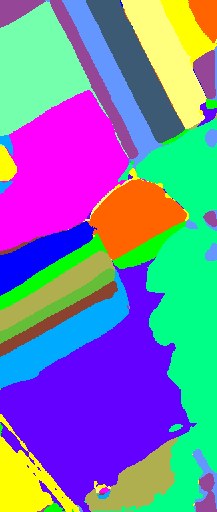}
    \centerline{(f) FreeNet}
  \end{minipage}
  \caption{Visualization of the classification maps for the Salinas dataset.
    (a) SVM. (b) S-CNN. (c) Gabor-CNN. (d) DFFN. (e) 3D-GAN. (f) FreeNet.}
  \label{fig:vis_salinas}
\end{figure*}

To further evaluate the effectiveness of the FPGA framework, we also performed experiments on the Salinas dataset.
The HSI of Salinas dataset has 512 $\times$ 217 pixels and 204 spectral bands with a spatial resolution of 3.7 m.
The ground-truth map covers 16 classes of interest.
Fig.~\ref{fig:salinas} shows the three-band false-color composite image and the corresponding ground-truth map.
The numbers of training and test samples are listed in Table.~\ref{tab:salinas_samples}.
The selection of samples was random selection with the same random seed as used in Experiment 1.

\begin{table}[htb]
  \caption{The number of training samples and test samples for the Salinas dataset
    \label{tab:salinas_samples}}
  \centering
  \renewcommand{\arraystretch}{1.5}
  \begin{tabular}{l|l|c|c|c}
    \hline
    Class & Class name                   & \#Training & \#Test & \#Total \\
    \hline
    C1    & Brocoli\_green\_weeds\_1     & 200        & 1809   & 2009    \\
    C2    & Brocoli\_green\_weeds\_2     & 200        & 3526   & 3726    \\
    C3    & Fallow                       & 200        & 1776   & 1976    \\
    C4    & Fallow\_rough\_plow          & 200        & 1194   & 1394    \\
    C5    & Fallow\_smooth               & 200        & 2478   & 2678    \\
    C6    & Stubble                      & 200        & 3759   & 3959    \\
    C7    & Celery                       & 200        & 3379   & 3579    \\
    C8    & Grapes\_untrained            & 200        & 11071  & 11271   \\
    C9    & Soil\_vinyard\_develop       & 200        & 6003   & 6203    \\
    C10   & Corn\_senesced\_green\_weeds & 200        & 3078   & 3278    \\
    C11   & Lettuce\_romaine\_4wk        & 200        & 868    & 1068    \\
    C12   & Lettuce\_romaine\_5wk        & 200        & 1727   & 1927    \\
    C13   & Lettuce\_romaine\_6wk        & 200        & 716    & 916     \\
    C14   & Lettuce\_romaine\_7wk        & 200        & 870    & 1070    \\
    C15   & Vinyard\_untrained           & 200        & 7068   & 7268    \\
    C16   & Vinyard\_vertical\_trellis   & 200        & 1607   & 1807    \\
    \hline
    Total & -                            & 3200       & 50929  & 54129   \\
    \hline
  \end{tabular}
\end{table}

Fig.~\ref{fig:vis_salinas} shows the visual performance of the different methods.
We can observe that the classification maps in Fig.~\ref{fig:vis_salinas}(d)-(f) (DFFN, 3D-GAN and FreeNet) are better than those in Fig.~\ref{fig:vis_salinas} (a)-(c).
Although DFFN and FreeNet obtain a similar accuracy, as shown in Table.~\ref{tab:result_salinas}, the result of FreeNet contains less noise in the classification map, achieving a better visual performance.
This suggests that the global spatial information is important for HSI classification.
The results of the CNN-based methods show the over-smoothing problem, whereas, surprisingly, the SVM method can obtain sharper edges.
We speculate that this is because these methods model the spatial context information.
This causes the label of a pixel to be not only dependent on the center pixel, but also the neighboring pixels.
Thus, the pixels near the edges of objects usually have different labels but a highly similar spatial context, which results in the over-smoothing problem.

\begin{table}[hbt]
  \caption{The classification results of SVM \cite{melgani2004classification}, S-CNN\cite{liu2017supervised}, Gabor-CNN \cite{chen2017hyperspectral}, DFFN \cite{song2018hyperspectral}, 3D-GAN \cite{zhu2018generative} and FreeNet on the Salinas dataset
    \label{tab:result_salinas}}
  \centering
  \renewcommand{\arraystretch}{1.5}
  \resizebox{\linewidth}{!}{
    \begin{tabular}{c|c|ccccc|c}
      \hline
      \multicolumn{2}{c|}{\multirow{2}{*}{Class}}    & \multicolumn{5}{c|}{Patch-based} & \multicolumn{1}{c}{Patch-free}                                        \\ \cline{3-8}
      \multicolumn{2}{c|}{}                          & SVM                              & S-CNN                          & Gabor-CNN & DFFN & 3D-GAN  & FreeNet         \\ \hline
      \multirow{16}{*}{\rotatebox{90}{Accuracy(\%)}} & C1                               & 99.61                          & 100       & 100    & 99.99 & 75.35 & 100   \\
                                                     & C2                               & 99.69                          & 97.75     & 88.06  & 99.94 & 98.49  & 100   \\
                                                     & C3                               & 99.56                          & 98.88     & 99.25  & 100  & 100  & 100   \\
                                                     & C4                               & 99.41                          & 100       & 100    & 100  &  99.76 & 99.92 \\
                                                     & C5                               & 98.72                          & 99.93     & 98.88  & 99.11 & 100 & 99.11 \\
                                                     & C6                               & 99.77                          & 89.48     & 100    & 99.95 & 99.97 & 100   \\
                                                     & C7                               & 99.52                          & 99.30     & 98.94  & 99.43 & 99.45  & 100   \\
                                                     & C8                               & 76.74                          & 98.69     & 99.48  & 99.56 & 98.30 & 99.94 \\
                                                     & C9                               & 99.37                          & 99.34     & 97.27  & 100  &  99.95 & 100   \\
                                                     & C10                              & 95.13                          & 100       & 99.21  & 99.79 & 99.79 & 99.77 \\
                                                     & C11                              & 99.32                          & 99.93     & 100    & 99.48 & 100 & 100   \\
                                                     & C12                              & 99.70                          & 99.89     & 100    & 99.84  & 100& 100   \\
                                                     & C13                              & 99.15                          & 88.62     & 100    & 99.96  & 100& 100   \\
                                                     & C14                              & 98.38                          & 88.72     & 99.79  & 99.95 & 100 & 100   \\
                                                     & C15                              & 75.56                          & 90.62     & 94.31  & 99.45 &99.52  & 99.99 \\
                                                     & C16                              & 99.23                          & 99.96     & 93.38  & 99.96 &90.25  & 99.88 \\ \hline
      \multicolumn{2}{c|}{OA(\%)}                    & 90.92                            & 97.62                          & 97.63     & 99.71& 98.22& 99.92           \\
      \multicolumn{2}{c|}{AA(\%)}                    & 96.18                            & 96.94                          & 98.04     & 99.78 & 97.75& 99.91           \\
      \multicolumn{2}{c|}{Kappa}                     & 0.8985                           & 0.9510                         & 0.9734    & 0.9967& 0.9793 & 0.9991          \\ \hline
    \end{tabular}
  }
\end{table}

Table.~\ref{tab:result_salinas} lists the accuracies of the patch-based methods and the patch-free method, where it can be seen that FreeNet performs much better than most of the patch-based methods in terms off OA, AA, and Kappa.
Compared to the state-of-the-art patch-based method of DFFN, FreeNet obtains a slightly higher OA of 99.92\%, exceeding DFFN by 0.2\%.
The proposed FreeNet achieves the highest accuracy among all the methods, slightly surpassing DFFN in all of the metrics.
FreeNet benefits from the proposed semantic-spatial feature fusion module (lateral based SSF), which is based on residual learning.
Furthermore, FreeNet fuses the features from the shallow layer in the encoder and the deep layer in the decoder to achieve the fusion of semantic information and spatial details.
It is interesting that the lateral based SSF in FreeNet and the multi-layer fusion in DFFN both belong to cross-layer feature fusion.
This indicates that cross-layer feature fusion is an important component of HSI classification.

\subsection{Experiment 3: CASI University of Houston Dataset}
\label{sec:exp2_grss2013}
The proposed  patch-free method achieved saturation on the ROSIS-03 Pavia University dataset and the Salinas dataset.
Although the proposed FreeNet surpasses the state-of-the-art methods on these two simple benchmark datasets, the performance difference is fairly small due to the saturation of the accuracy.
Therefore, we chose the CASI University of Houston dataset for Experiment 3, which is a relatively difficult benchmark dataset, to clearly present the performance difference.

\begin{figure}[htb]
  \begin{minipage}[b]{\linewidth}
    \centering
    \includegraphics[width=\linewidth]{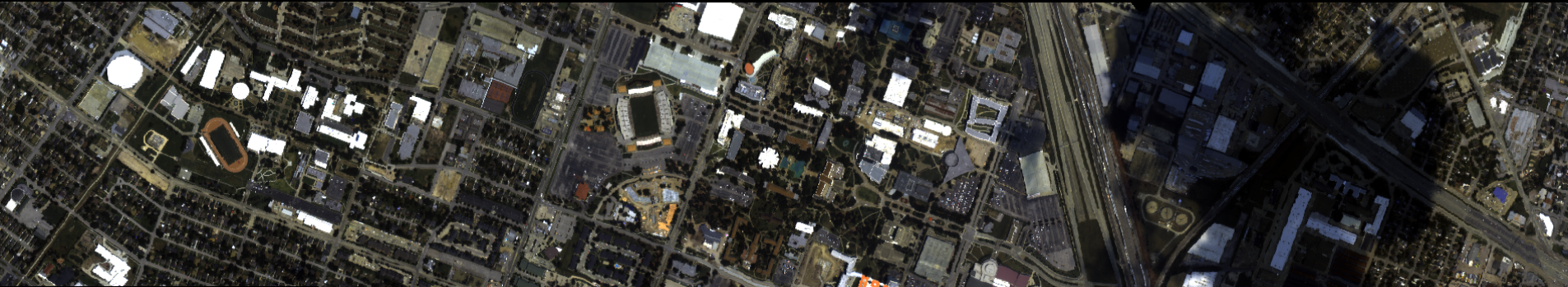}
    \centerline{(a) Hyperspectral image}
  \end{minipage}
  \vfill
  \vspace{4pt}
  \begin{minipage}[b]{\linewidth}
    \centering
    \includegraphics[width=\linewidth]{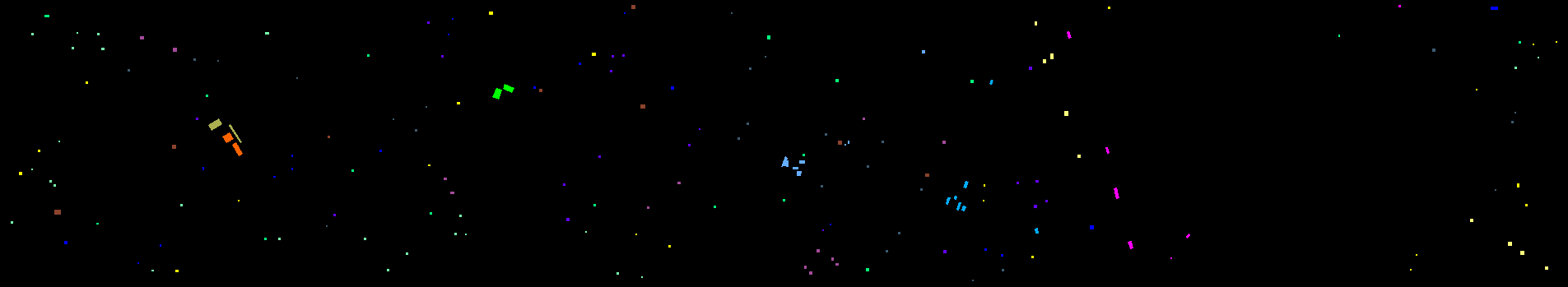}
    \centerline{(b) Training samples}
  \end{minipage}
  \vfill
  \vspace{4pt}
  \begin{minipage}[b]{\linewidth}
    \centering
    \includegraphics[width=\linewidth]{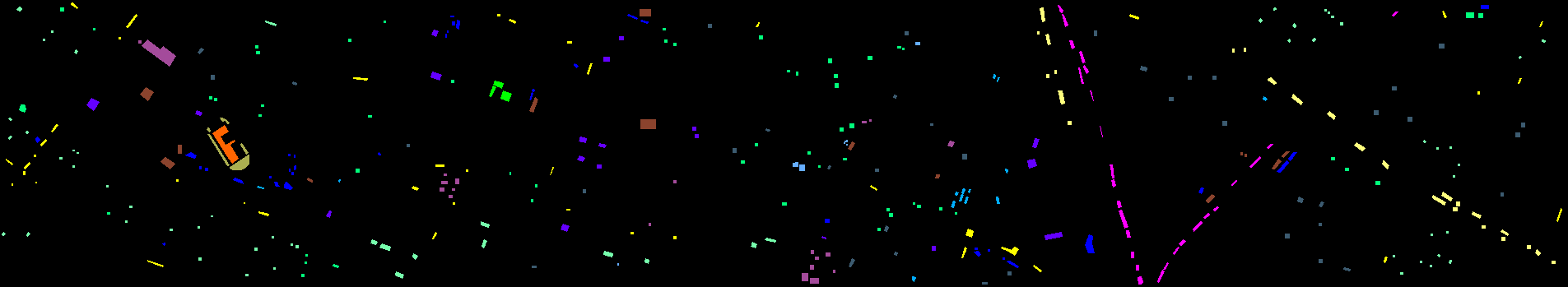}
    \centerline{(c) Test samples}
  \end{minipage}
  \vfill
  \vspace{4pt}
  \begin{minipage}[b]{\linewidth}
    \centering
    \includegraphics[width=\linewidth]{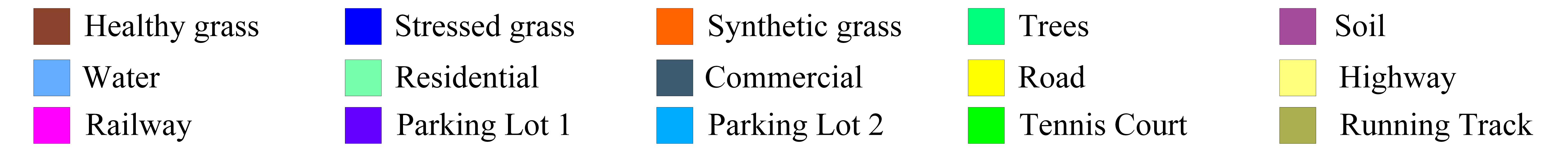}
    \centerline{(d) Legend}
  \end{minipage}
  \caption{The CASI University of Houston data set. (a) Color composite representation of the hyperspectral data using bands of 70, 50, and 20, as red, green, and blue, respectively; (b) Training samples; (c) Test samples; (d) Legend}
  \label{fig:grss2013}
\end{figure}

The CASI University of Houston dataset was published as part of the 2013 IEEE Geoscience and Remote Sensing Society (GRSS) data fusion contest.
The HSI has 349 $\times$ 1905 pixels with 144 spectral bands and a spatial resolution of 2.5 m, the wavelength of which ranges from 0.38 to \SI{1.05}{\micro\metre}.

Fig.~\ref{fig:grss2013} (a) shows the false-color composite of the hyperspectral image.
The dataset provides a ground-truth map of 15 classes, as shown in Fig.~\ref{fig:grss2013} (b) and (c).
The corresponding legend is shown in Fig.~\ref{fig:grss2013} (d).
Table.~\ref{tab:grss2013_samples} lists the number of training and test samples per class.
Differing from the first two datasets, the CASI University of Houston dataset provides officially predefined training and test samples.
Thus, the results obtained with this benchmark dataset can be considered as more reliable and stable.

\begin{figure}[htb]
  \begin{minipage}[b]{\linewidth}
    \centering
    \includegraphics[width=\linewidth]{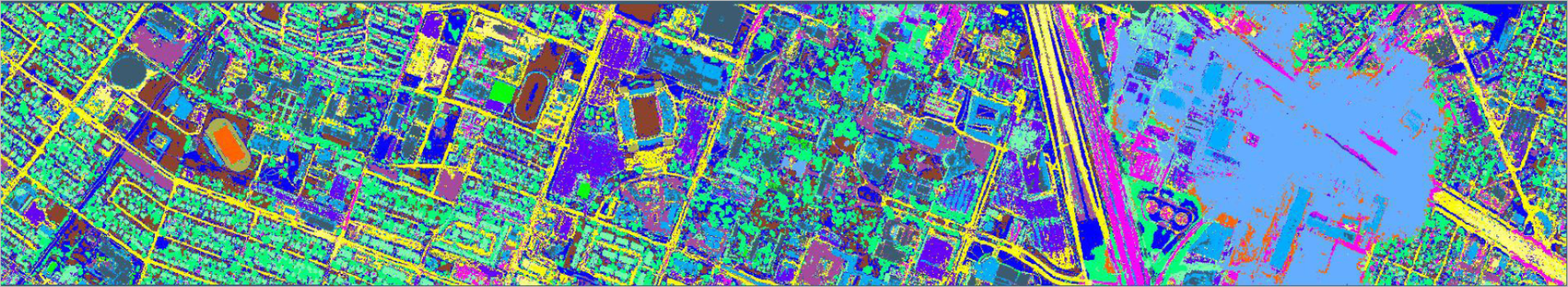}
    \centerline{(a) SVM}
  \end{minipage}
  \vfill
  \vspace{4pt}
  \begin{minipage}[b]{\linewidth}
    \centering
    \includegraphics[width=\linewidth]{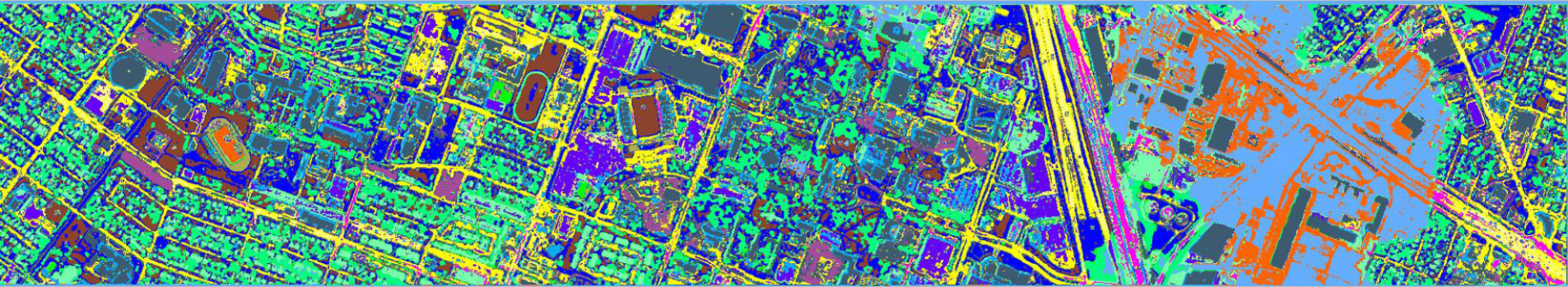}
    \centerline{(b) S-CNN}
  \end{minipage}
  \vfill
  \vspace{4pt}
  \begin{minipage}[b]{\linewidth}
    \centering
    \includegraphics[width=\linewidth]{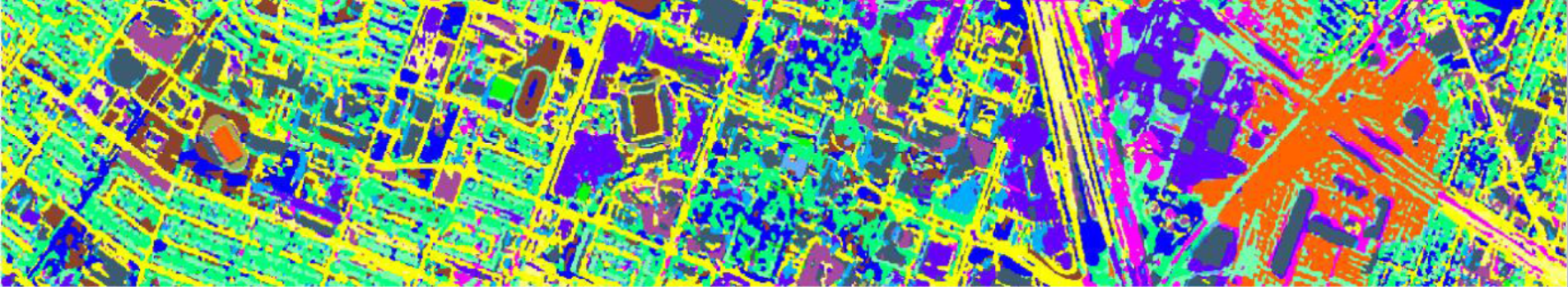}
    \centerline{(c) Gabor-CNN}
  \end{minipage}
  \vfill
  \vspace{4pt}
  \begin{minipage}[b]{\linewidth}
    \centering
    \includegraphics[width=\linewidth]{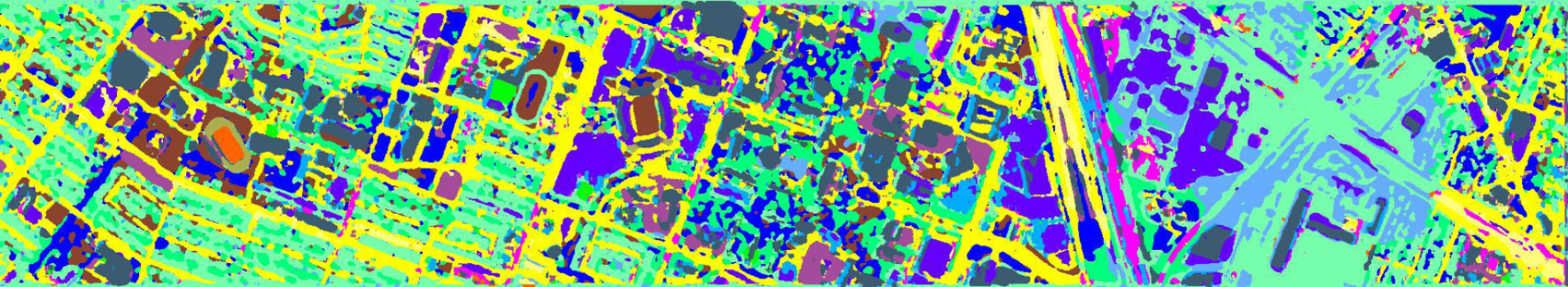}
    \centerline{(d) DFFN}
  \end{minipage}
  \vfill
  \vspace{4pt}
  \begin{minipage}[b]{\linewidth}
    \centering
    \includegraphics[width=\linewidth]{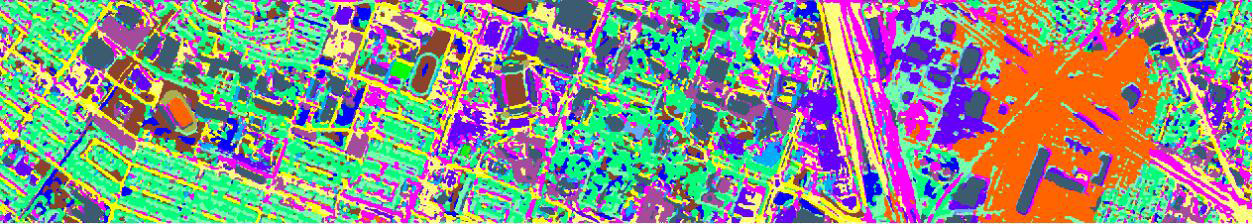}
    \centerline{(e) 3D-GAN}
  \end{minipage}
  \vfill
  \vspace{4pt}
  \begin{minipage}[b]{\linewidth}
    \centering
    \includegraphics[width=\linewidth]{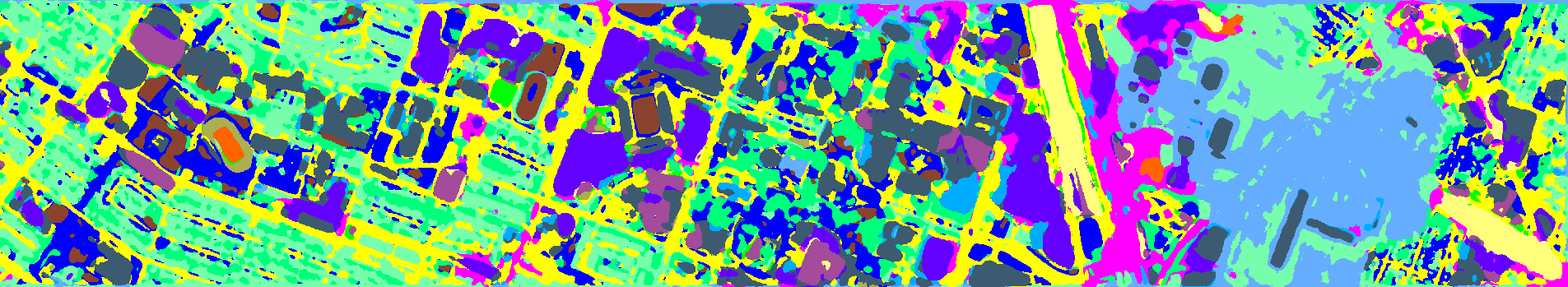}
    \centerline{(f) FreeNet}
  \end{minipage}
  \caption{Visualization of the classification maps for the CASI University of Houston dataset.
    (a) SVM. (b) S-CNN. (c) Gabor-CNN. (d) DFFN. (e) 3D-GAN. (f) FreeNet.}
  \label{fig:vis_grss2013}
\end{figure}

\begin{table}[htb]
  \caption{The numbers of training samples and test samples for the CASI University of Houston dataset
    \label{tab:grss2013_samples}}
  \centering
  \renewcommand{\arraystretch}{1.5}
  \begin{tabular}{l|l|c|c|c}
    \hline
    Class & Class name      & \#Training & \#Test & \#Total \\
    \hline
    C1    & Grass–Healthy   & 198        & 1053   & 1251    \\
    C2    & Grass–Stressed  & 190        & 1064   & 1254    \\
    C3    & Grass–Synthetic & 192        & 505    & 697     \\
    C4    & Tree            & 188        & 1056   & 1244    \\
    C5    & Soil            & 186        & 1056   & 1242    \\
    C6    & Water           & 182        & 143    & 325     \\
    C7    & Residential     & 196        & 1072   & 1268    \\
    C8    & Commercial      & 191        & 1053   & 1244    \\
    C9    & Road            & 193        & 1059   & 1252    \\
    C10   & Highway         & 191        & 1036   & 1227    \\
    C11   & Railway         & 181        & 1054   & 1235    \\
    C12   & Parking Lot 1   & 192        & 1041   & 1234    \\
    C13   & Parking Lot 2   & 184        & 285    & 469     \\
    C14   & Tennis Court    & 181        & 247    & 428     \\
    C15   & Running Track   & 187        & 473    & 660     \\
    \hline
    Total & -               & 2832       & 12179  & 15011   \\
    \hline
  \end{tabular}
\end{table}

\begin{table}[h]
  \caption{The classification results of SVM \cite{melgani2004classification}, S-CNN\cite{liu2017supervised}, Gabor-CNN \cite{chen2017hyperspectral}, DFFN \cite{song2018hyperspectral}, 3D-GAN \cite{zhu2018generative} and FreeNet on the CASI University of Houston dataset.
    \label{tab:result_grss2013}}
  \centering
  \renewcommand{\arraystretch}{1.5}
  \resizebox{\linewidth}{!}{
    \begin{tabular}{c|c|ccccc|c}
      \hline
      \multicolumn{2}{c|}{\multirow{2}{*}{Class}}    & \multicolumn{5}{c|}{Patch-based} & \multicolumn{1}{c}{Patch-free}                                        \\ \cline{3-8}
      \multicolumn{2}{c|}{}                          & SVM                              & S-CNN                          & Gabor-CNN & DFFN  & 3D-GAN & FreeNet         \\ \hline
      \multirow{15}{*}{\rotatebox{90}{Accuracy(\%)}} & C1                               & 82.05                          & 83.00     & 82.30  & 77.41 & 81.58  & 80.91 \\
                                                     & C2                               & 80.55                          & 83.27     & 84.24  & 81.39 & 79.74 & 84.21 \\
                                                     & C3                               & 100                            & 98.66     & 95.61  & 94.59 & 97.42 & 98.02 \\
                                                     & C4                               & 92.52                          & 93.50     & 92.39  & 87.82 & 93.36 & 91.95 \\
                                                     & C5                               & 98.11                          & 96.91     & 99.80  & 96.05 & 99.71 & 100   \\
                                                     & C6                               & 95.10                          & 94.12     & 96.47  & 96.15 & 95.08 & 96.50 \\
                                                     & C7                               & 75.00                          & 79.95     & 84.58  & 80.25 & 89.90 & 88.53 \\
                                                     & C8                               & 40.17                          & 68.19     & 73.09  & 77.78 & 70.52 & 74.83 \\
                                                     & C9                               & 74.88                          & 76.39     & 78.14  & 84.66 & 54.89 & 87.72 \\
                                                     & C10                              & 51.64                          & 48.10     & 58.01  & 64.63 & 49.89 & 62.25 \\
                                                     & C11                              & 78.37                          & 74.64     & 73.35  & 88.61 & 77.36 & 83.40 \\
                                                     & C12                              & 68.40                          & 85.49     & 86.74  & 98.57 & 60.46 & 98.84 \\
                                                     & C13                              & 69.47                          & 88.58     & 91.16  & 83.09 & 81.71 & 88.42 \\
                                                     & C14                              & 100                            & 99.19     & 100    & 99.72 & 95.14 & 96.76 \\
                                                     & C15                              & 98.10                          & 96.40     & 77.73  & 81.90 & 65.35  & 94.29 \\ \hline
      \multicolumn{2}{c|}{OA(\%)}                    & 76.88                            & 82.34                          & 84.32     & 84.56 & 78.16 & 86.61           \\
      \multicolumn{2}{c|}{AA(\%)}                    & 80.29                            & 84.43                          & 84.17     & 86.18  & 79.98 & 88.44           \\
      \multicolumn{2}{c|}{Kappa}                     & 0.7513                           & 0.8052                         & 0.8114    & 0.8328 & 0.7616 & 0.8555          \\ \hline
    \end{tabular}
  }
\end{table}

Fig.~\ref{fig:vis_grss2013} shows the classification map of the compared methods for the visual performance estimation.
It can be clearly observed that the maps in Fig.~\ref{fig:vis_grss2013} (c)-(f) are clearer and contain less noise than those in Fig.~\ref{fig:vis_grss2013} (a)-(b).
For the Road, Highway, and Railway classes, these three classes in the classification map of FreeNet show better connectivity.
Meanwhile, the accuracy for these three classes is higher than for the other methods as shown in Table.~\ref{tab:result_grss2013} C9-C11.
We speculate that the increased spatial context of these classes enhances the discriminative ability for each class.

\begin{figure}[hbt]
  \centering
  \includegraphics[width=\linewidth]{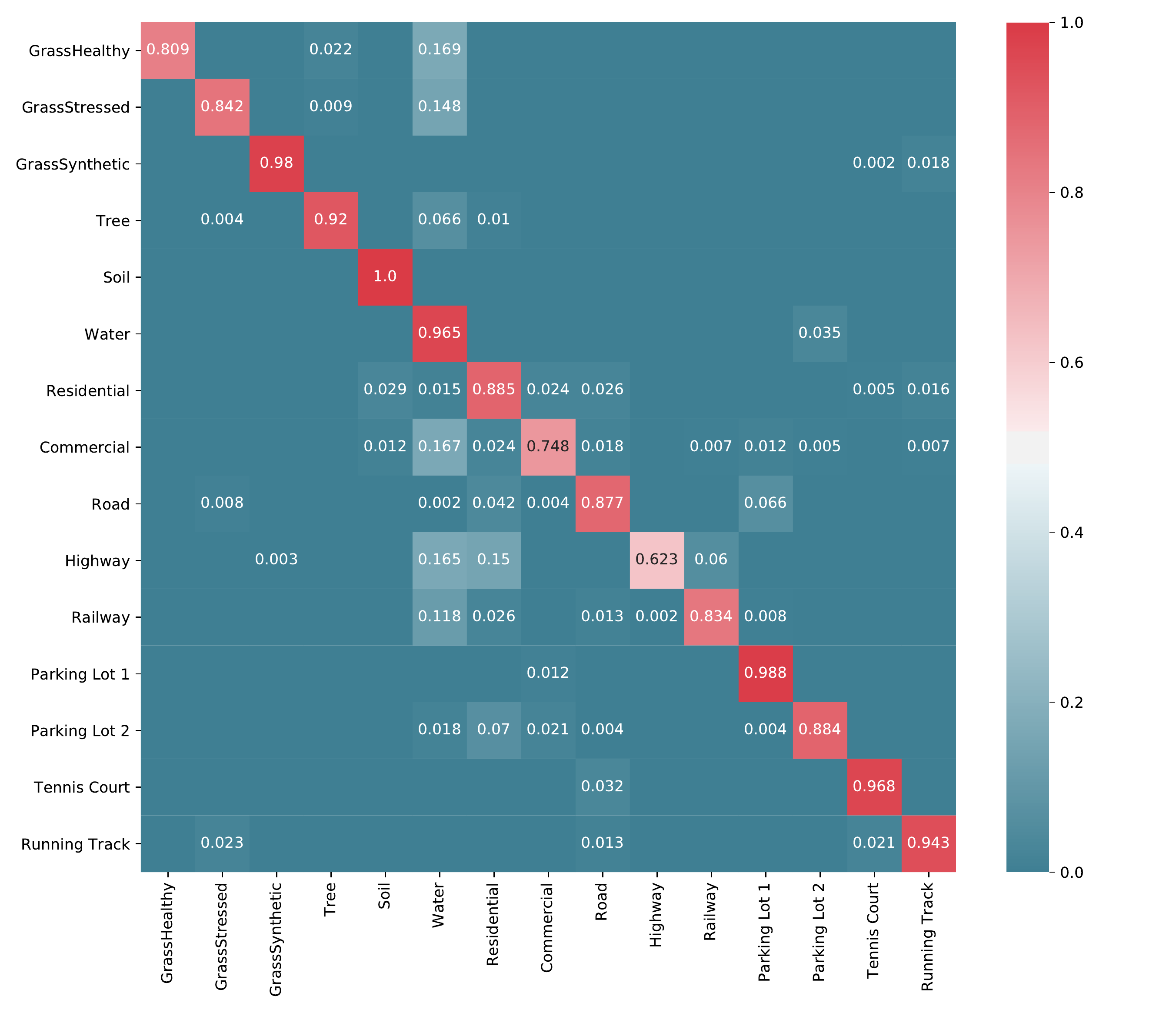}
  \caption{Confusion matrix for FreeNet on the CASI University of Houston dataset.}
  \label{fig:cm_freenet_grss2013}
\end{figure}

Table.~\ref{tab:result_grss2013} lists the classification results obtained on this dataset.
FreeNet achieves a state-of-the-art result and achieves a $\sim$2\% improvement over the best result (DFFN) of the patch-based methods.
This suggests that the patch-free global learning framework performs even better than the patch-based local learning framework.
FreeNet benefits from the increased global spatial information, including the global spatial context and spatial details.
By replacing the patch with a larger receptive field, the model under the patch-free global learning framework is able to obtain more global spatial context, which boosts the classification performance.
FreeNet makes full use of the global spatial context via the spectral attention module and the lateral connection based SSF.
The spectral attention module utilizes the global spatial context embedding vector to re-weight the feature maps for modeling the interdependencies between feature maps, which aims to estimate the importance of the different feature maps by the global spatial context.
This benefits the classification of HSIs with redundant spectral information.
Furthermore, the lateral connection based SSF leverages the finer spatial detail features to refine semantically stronger but spatially coarser deep features, which are aggregated to enhanced features with finer spatial details and stronger semantic information.

The accuracy obtained on this dataset is clearly lower than the accuracies obtained on the other two datasets using the same method.
To explore the reason for this, we drew the confusion matrix for FreeNet on this dataset for  a quantitative analysis, as shown in Fig.~\ref{fig:cm_freenet_grss2013}.
We can see that the other categories tend to be wrongly classified as Water.
For example, a large amount of GrassHealthy, Commercial, and Railway pixels are wrongly classified as Water. 
This is because these three categories are distributed in the shadow area of the HSI, as shown in Fig.~\ref{fig:grss2013} (a).
The shadow significantly influences the spectral information, which misleads the classification.
Qualitatively, from the visualization result (Fig.~\ref{fig:vis_grss2013} (e)), the shadow area is occupied by water.
This indicates that the spectral information is sensitive to the observation conditions, which haves a great impact on the qualitative and quantitative performance of HSI classification.

\begin{table*}[ht]
  \caption{
    HSI classification results evaluated on the CASI University of Houston dataset. Starting from our encoder-decoder baseline, the GS$^2$ sampling strategy, lateral connection and spectral attention are gradually added in FreeNet for the module analysis.
    SA denotes spectral attention and LC denotes lateral connection based SSF.
    \label{tab:ablation_study}}
  \centering
  \renewcommand{\arraystretch}{1.5}
  \begin{tabular}{c|l|ccc|ccc}
    \hline
    Compression factor            & Method                    & GS$^2$ sampling strategy & Lateral connection & Spectral attention & OA    & AA    & Kappa  \\ \hline
    \multirow{4}{*}{$\beta=0.75$} & (a) Baseline              & -                        & -                  & -                  & 15.23 & 19.54 & 0.0972 \\ \cline{2-8}
                                  & (b) FreeNet w/o LC and SA & $\checkmark$             &                    &                    & 65.12 & 67.45 & 0.6225 \\
                                  & (c) FreeNet w/o SA        & $\checkmark$             & $\checkmark$       &                    & 84.23 & 85.61 & 0.8293 \\
                                  & (d) FreeNet               & $\checkmark$             & $\checkmark$       & $\checkmark$       & 85.49 & 86.64 & 0.8423 \\ \hline
    \multirow{4}{*}{$\beta=1.0$}  & (a) Baseline              & -                        & -                  & -                  & 12.49 & 13.89 & 0.0657 \\ \cline{2-8}
                                  & (b) FreeNet w/o LC and SA & $\checkmark$             &                    &                    & 65.16 & 65.2  & 0.6212 \\
                                  & (c) FreeNet w/o SA        & $\checkmark$             & $\checkmark$       &                    & 84.91 & 86.25 & 0.8363 \\
                                  & (d) FreeNet               & $\checkmark$             & $\checkmark$       & $\checkmark$       & 86.61 & 88.44 & 0.8555 \\ \hline
  \end{tabular}
\end{table*}

\begin{figure*}[htb]
  \begin{minipage}[b]{.32\linewidth}
    \centering
    \includegraphics[width=\linewidth]{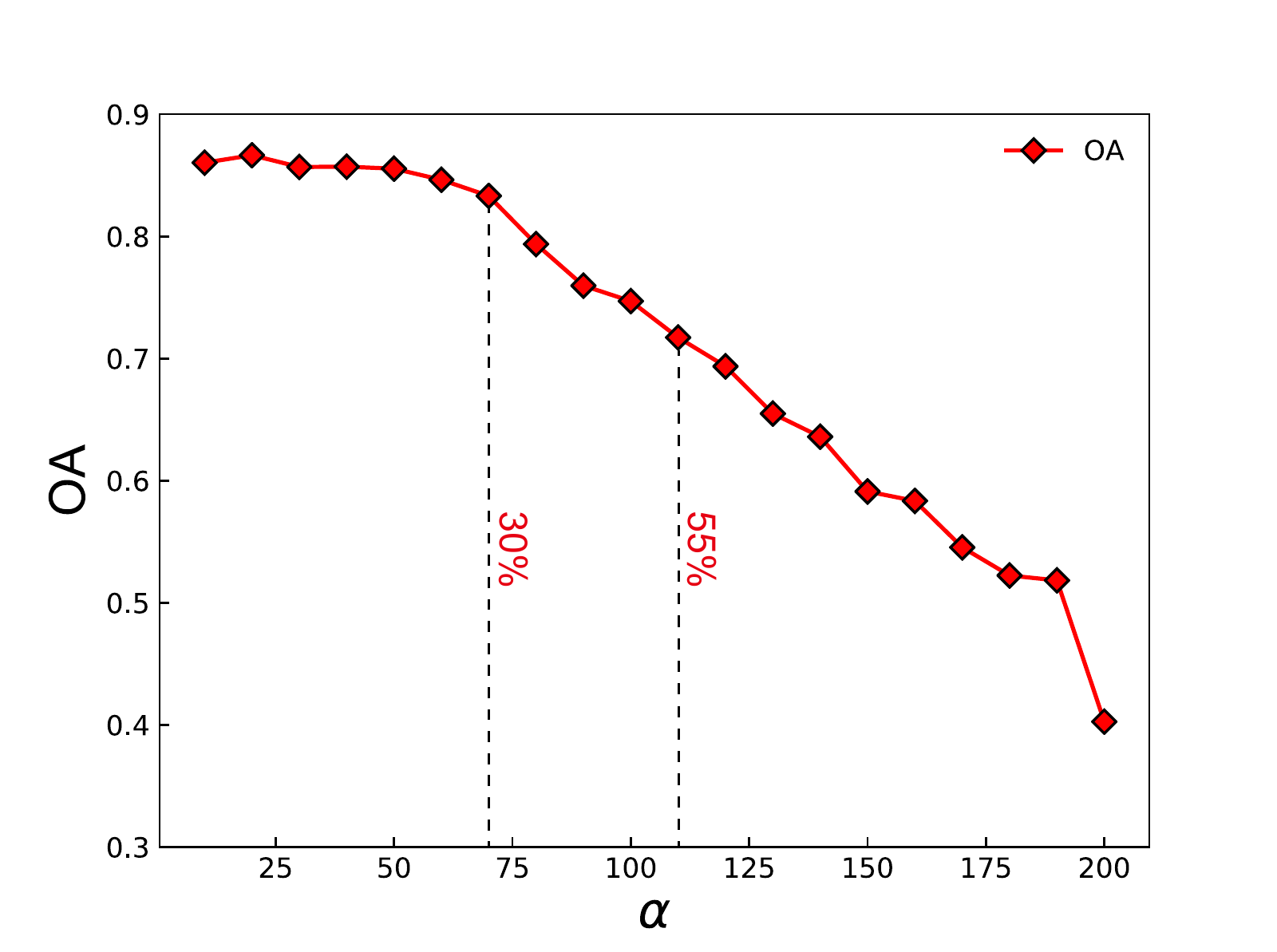}
    \centerline{(a)}
  \end{minipage}
  \hfill
  \begin{minipage}[b]{0.32\linewidth}
    \centering
    \includegraphics[width=\linewidth]{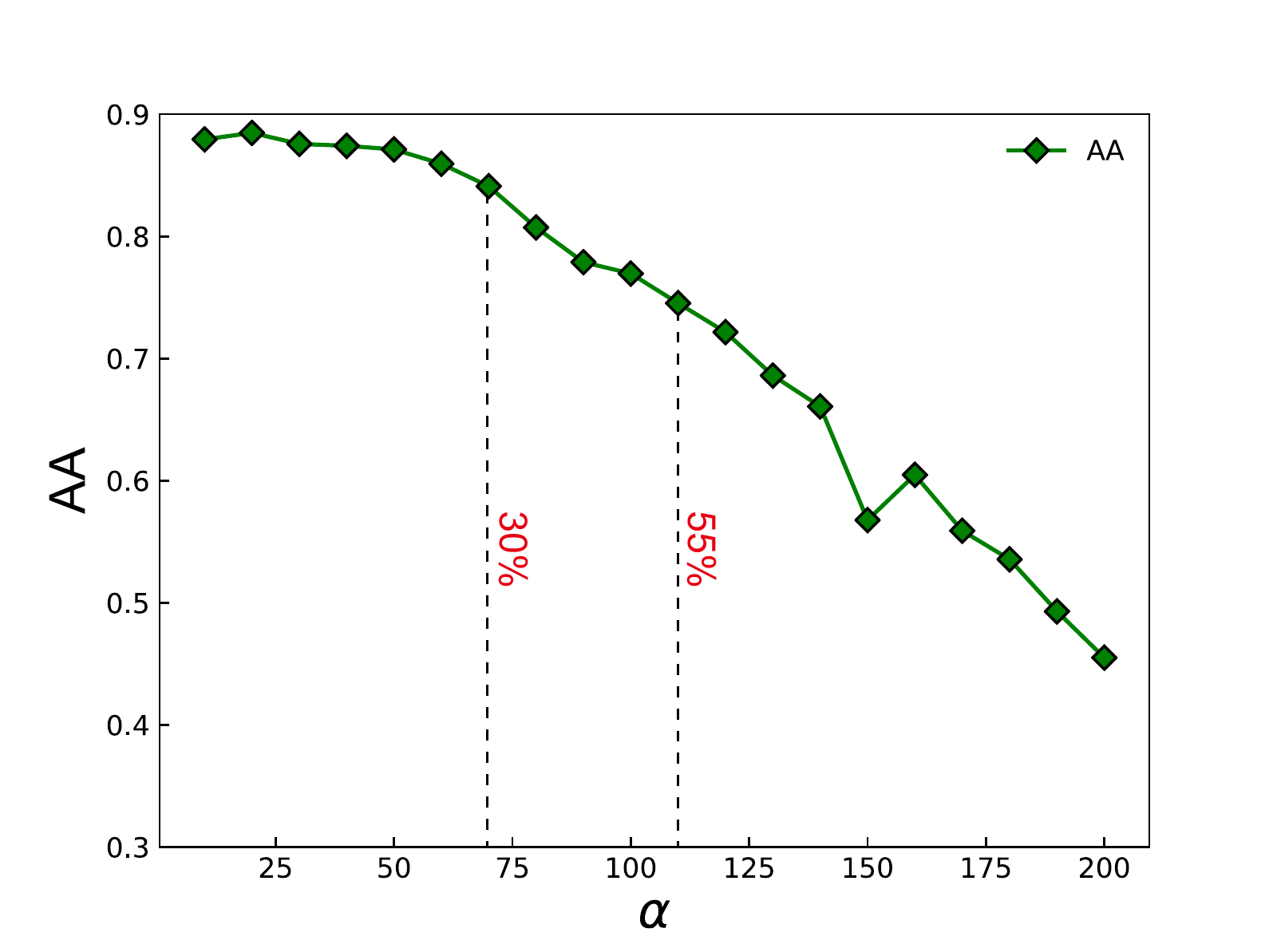}
    \centerline{(b)}
  \end{minipage}
  \hfill
  \begin{minipage}[b]{0.32\linewidth}
    \centering
    \includegraphics[width=\linewidth]{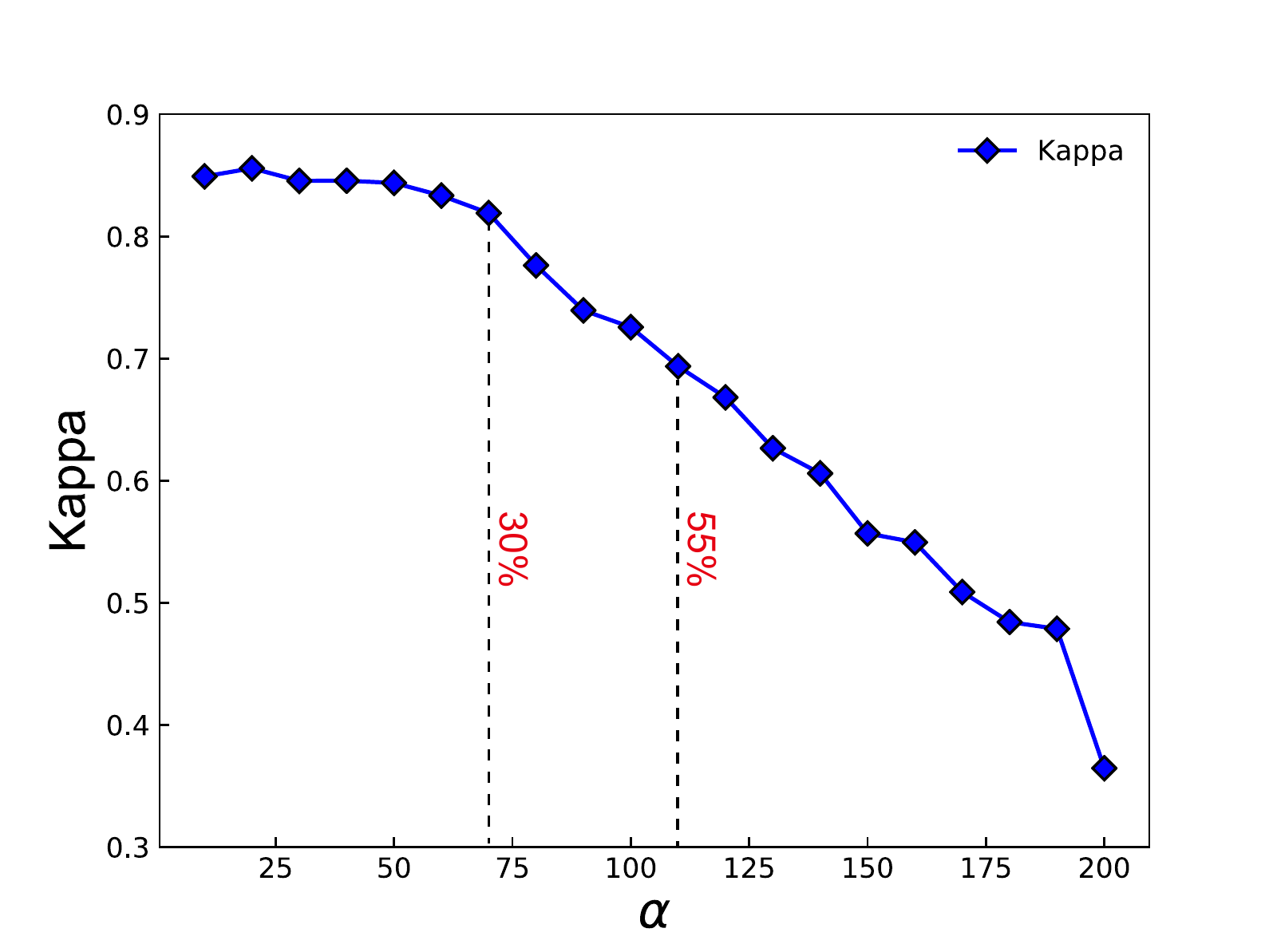}
    \centerline{(c)}
  \end{minipage}
  \caption{Sensitivity of the mini-batch size per class ($\alpha$) in the GS$^2$ sampling strategy on the CASI University of Houston dataset.
    (a) The impact on the OA of different $\alpha$ settings.
    (b) the impact on the AA of different $\alpha$ settings.
    (c) the impact on the Kappa of different $\alpha$ settings.}
  \label{fig:alpha}
\end{figure*}

% ----------------------------------------------------------------

\section{FPGA Sensitivity Analysis}
To clearly understand the effectiveness of each component in the proposed FreeNet under the patch-free global learning framework, we conducted extensive module analysis experiments.
All the component analysis experiments were performed on the CASI University of Houston dataset as this dataset has official training and test samples.
We chose the CASI University of Houston dataset for more stable, reliable and reproducible analysis results.
The baseline methods shown in Table.\ref{tab:ablation_study} (a) are encoder-decoder FCNs with $\beta=0.75$ and $1.0$, respectively, which were trained by directly using all the training samples, without any sampling strategy.
%---------------------------------------------------------
\subsection{\textbf{The GS$^2$ Sampling Strategy}}
\label{sec:gs2}
Table.~\ref{tab:ablation_study} (b) presents the results of the baseline methods with the GS$^2$ sampling strategy.
The results indicate that the GS$^2$ sampling strategy improves the OA for both $\beta=0.75$ (from 15.23\% to 65.12\%) and $\beta=1.0$ (from 12.49\% to 65.16\%).
A model with an OA of $\sim$10\% on this dataset means that this model fails to converge.
The GS$^2$ sampling strategy effectively addresses this issue for the baseline encoder-decoder FCN.
The GS$^2$ sampling strategy splits the original training samples into many mini-batch training samples with the same spatial size to obtain diverse gradients and partially supervised signals.
These diverse gradients and partially supervised signals make it easier to skip local minimum points.
This suggests that it is very important for the end-to-end trainable FCNs used in HSI classification to obtain more diverse gradients.

The hyperparameter $\alpha$ is introduced in the GS$^2$ sampling strategy to control the number of samples in the mini-batch per class.
The classification results of FreeNet ($\beta=1.0$) with different value of $\alpha$ from 10 to 200 with an interval of 10 are shown in Fig.~\ref{fig:alpha}.
We can observe that the proposed FreeNet shows a stable performance when $\alpha$ is set to a value within 30\% of the number of total training samples.
This indicates the robustness of the GS$^2$ sampling strategy with respect to $\alpha$ when the stochasticity is sufficient.
When $\alpha$ is set to a value larger than 55\% of the number of total training samples, the classification accuracy of FreeNet drops rapidly, to even lower than the SVM baseline.
The reason for this is that $\alpha$ indirectly controls the stochasticity of the sampling.
A smaller $\alpha$ means greater stochasticity of the sampling.
Meanwhile, the stochasticity of the sampling directly influences the diversity of the gradients.
Therefore, FreeNet with a smaller $\alpha$ always obtains a higher accuracy, while a larger $\alpha$ brings a worse performance.
%---------------------------------------------------------
\subsection{\textbf{Lateral Connection Based SSF}}
Lateral connection based SSF is described in Section \ref{sec:ssf}.
Table.~\ref{tab:ablation_study} (c) presents the effectiveness of the lateral connection based SSF.
When applying FreeNet without the lateral connection (LC) and spectral attention (SA) modules (Table.~\ref{tab:ablation_study} (b)), the classification performance shows a significant reduction.
The addition of the lateral connection and spectral attention modules to FreeNet ($\beta=0.75$) results in an OA improvement from 65.12\% to 84.23\%, and The addition of the lateral connection and spectral attention modules to FreeNet ($\beta=1.0$) results in an OA improvement from 65.16\% to 84.91\%, achieving a similar performance to the state-of-the-art patch-based HSI classifiers.
With the help of the lateral connection based SSF, the spatial detail features of the shallow convolutional layers can be passed on to the decoder to progressively refine the spatial detail of the deep semantic features, thus obtaining spatially finer and semantically stronger fused features.
Meanwhile, the aggregation function of pointwise addition can alleviate the gradient vanishing problem, making the optimization easier.
This suggests that lateral connection based SSF is important for HSI classification when using an encoder-decoder architecture under the patch-free global learning framework.

% ---------------------------------------------------------
\subsection{\textbf{Spectral Attention}}
Table.~\ref{tab:ablation_study} (d) presents the effectiveness of the spectral attention module.
Based on FreeNet without the spectral attention module (Table.~\ref{tab:ablation_study} (c)), the spectral attention module brings an additional improvement to FreeNet ($\beta=0.75$) (84.23\% to 85.49) and FreeNet ($\beta=1.0$) (84.91\% to 86.61\%).
The spectral attention module models the interdependencies of the feature maps in the encoder of FreeNet via the global spatial context, boosting the classification performance.
This indicates that it is valuable to further exploit the raw redundant spectral features guided by spatial context for HSI classification.

% ----------------------------------------------------------------
\subsection{\textbf{Model Complexity and Inference Speed Analysis}}
For a fair comparison of the model complexity and inference speed between the patch-based and patch-free methods, we directly used the encoder of FreeNet followed by a $1\times 1$ convolutional layer as the classifier, to represent the patch-based method.
Three variants ($\beta=0.5, 0.75$ and $1.0$) of FreeNet and the corresponding encoders were used as a comparison.
During the inference, the batch size of the patch-based methods was set to 1024 for parallel computation.
The image used for the benchmarking was from the CASI University of Houston dataset, which was converted to a 3-D float32 tensor of shape [349, 1905, 144].

We adopted the number of parameters (\# params) to measure the model complexity and the giga floating-point operations per second (GFLOPs) to measure the theoretical computational overhead.
Meanwhile, the GPU time cost was used to measure the actual efficiency of the models.

\begin{table}[hbt]
  \caption{
    The model complexity and inference speed of patch-based and patch-free methods.
    Encoder ($\beta$) is simply implemented by encoder in corresponding FreeNet with same $\beta$ and a $1\times 1$ convolutional layer.
    \label{tab:speed}}
  \centering
  \renewcommand{\arraystretch}{1.5}
  \begin{tabular}{l|c|cc}
    \hline
    \multirow{2}{*}{Methods}          & Model complexity      & \multicolumn{2}{c}{Inference speed}                        \\ \cline{2-4}
                                      & \#Params (M)          & GFLOPs                              & GPU (s)              \\ \hline
    \hspace{-5pt}\textit{Patch-based} & \multicolumn{1}{l|}{} & \multicolumn{1}{l}{}                & \multicolumn{1}{l}{} \\
    Encoder ($\beta=0.5$)             & 0.554                 & \multicolumn{1}{r}{53644.8}         & 55.221               \\
    Encoder ($\beta=0.75$)            & 1.191                 & \multicolumn{1}{r}{107289.6}        & 69.319               \\
    Encoder ($\beta=1.0$)             & 2.070                 & \multicolumn{1}{r}{167640.0}        & 84.106               \\ \hline
    \hspace{-5pt}\textit{Patch-free}  &                       &                                     &                      \\
    FreeNet ($\beta=0.5$)             & 0.724                 & 112.37                              & 0.094                \\
    FreeNet ($\beta=0.75$)            & 1.575                 & 220.82                              & 0.122                \\
    FreeNet ($\beta=1.0$)             & 2.749                 & 364.11                              & 0.146                \\ \hline
  \end{tabular}
\end{table}

\begin{figure*}[ht]
  \centering
  \includegraphics[width=\linewidth]{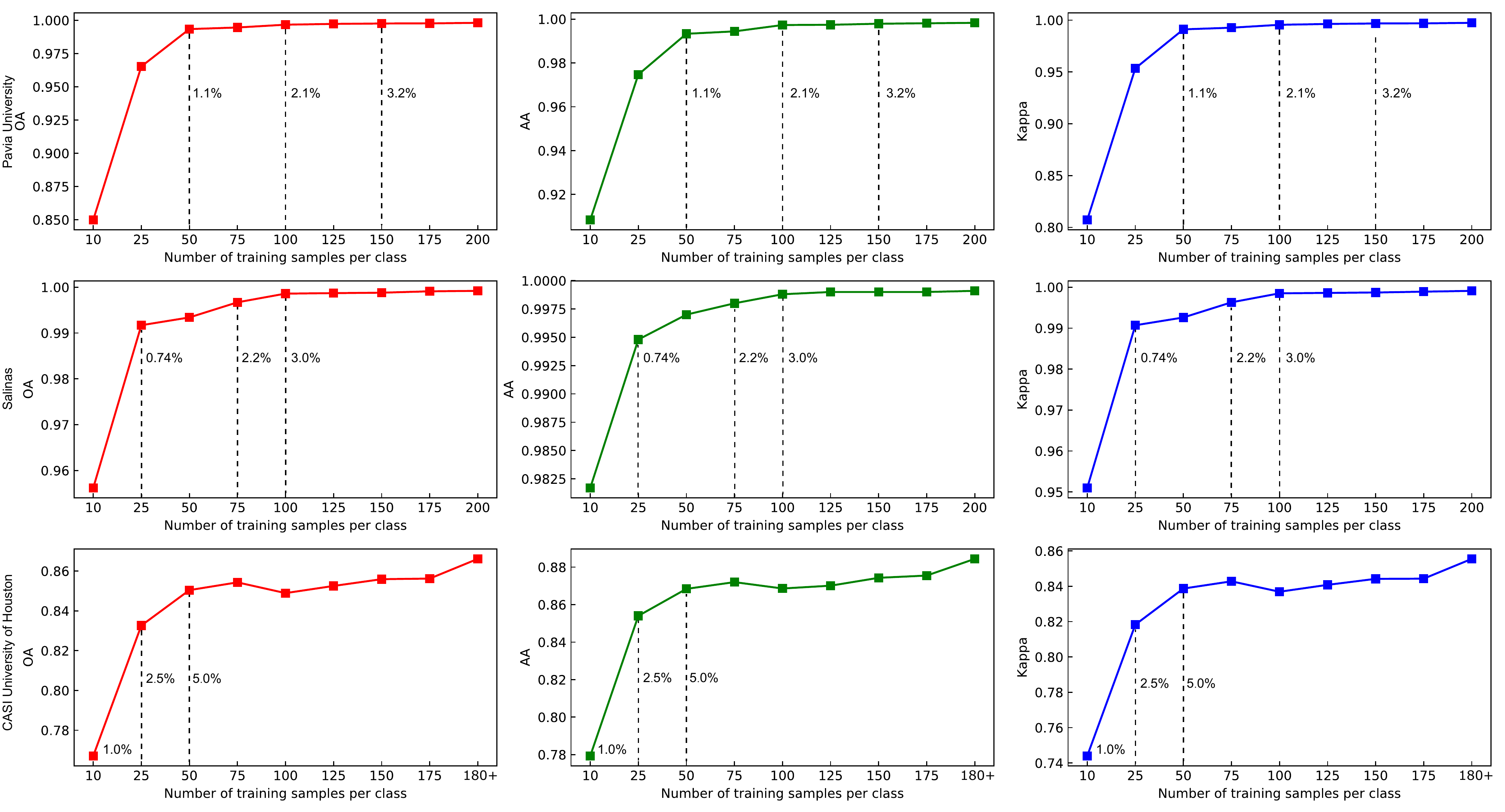}
  \caption{Performance versus the number of training samples per class. 
  These three rows represent the performance on Pavia University, Salinas and CASI University of Houston dataset, respectively. 
  The percentage near the dashed line is the percentage of training samples.}
  \label{fig:num_training_samples}
\end{figure*}

Table.~\ref{tab:speed} lists the measured results.
The results suggest that the patch-free methods are much faster than the patch-based methods, in both theory and practice, by avoiding redundant computation on the overlapping area.
Note that the practical speedup ratio ($\sim$560) is always larger than the theoretical speedup ratio ($\sim$480) because we ignore the influence of the parallel computation.
Although the patch-based methods have fewer parameters than the patch-free methods, the actual computation of the patch-based methods is slower than that of the patch-free methods.
Therefore, for real-time applications, such as HSI classification on unmanned aerial vehicle (UAV) or satellite imagery, the patch-free method is a better proposal than the patch-based method.

We also list the detailed running time costs of FreeNet for each dataset in Table.~\ref{tab:runingtime}.
It can be seen that the inference time cost is much smaller than the training time cost for each HSI.
This suggests that FreeNet is suitable for the application scenarios which allow offline learning and online inference, such as training FreeNet on ground station data and performing inference on UAV or satellite data.  

\begin{table}[]
  \caption{
    Detailed running time costs of FreeNet
    \label{tab:runingtime}}
  \centering
  \renewcommand{\arraystretch}{1.5}
  \begin{tabular}{l|c|c}
    \hline
    Dataset                    & Training time (s) & Test time (s) \\ \hline
    ROSIS-03 Pavia University  & 202               & 0.039         \\
    Salinas                    & 158               & 0.030         \\
    CASI University of Houston & 523               & 0.146         \\ \hline
  \end{tabular}
\end{table}

\subsection{\textbf{The Impact of the Number of Training Samples}}
To study the impact of the number of training samples on FreeNet, we conducted extensive experiments on the Pavia University, Salinas, and CASI University of Houston benchmark datasets.
The results are plotted in Fig.~\ref{fig:num_training_samples}.
Overall, reducing the training samples results in a drop in the performance of FreeNet. 
For the Pavia University dataset and Salinas dataset, the decreases in OA, AA, and Kappa are relatively small when using 1\% $\sim$ 3\% of the labeled samples.
It suggests that FreeNet is also robust, only using limited training sample number.
In order to further explore the effectiveness of FreeNet, we also trained FreeNet using only 10 samples per class. 
Compared with the Salinas dataset, the decrease in performance on the Pavia dataset is much more significant, at 14.82\% of OA, 8.99\% of AA, and 0.19 of Kappa. 
Meanwhile the decrease in accuracy for the Salinas dataset is 4.3\% of OA, 1.74\% of AA, and 0.048 of Kappa. 
This indicates that classification on the Salinas dataset with FreeNet is easier than for the Pavia University dataset 
We speculate that FreeNet benefits from more spectral information via the spectral attention module. 
This is because the spatial resolution of these two datasets is high, so that the contribution of the spatial information to the performance has achieved saturation. 
However the Salinas dataset has more bands, which may be the core reason for the easier classification. For the CASI University of Houston dataset, the performance is steady until the training samples are reduced to 5\% of the labeled samples. 
This dataset has fewer labeled samples than the other two datasets. 
Thus, the entries with a high percentage still have minimal training samples, which causes the performance to drop rapidly.

\section{Conclusion}
\label{sec:conclusion}
In this paper, we have proposed a fast patch-free global learning (FPGA) framework, pushing HSI classification further on both speed and accuracy.
In the FPGA framework, the GS$^2$ sampling strategy is proposed to ensure encoder-decoder based FCN training convergence by transforming the entire training samples into a stochastic sequence of class-stratified samples, to obtain stable and diverse gradients.
After ensuring the convergence of the training, FreeNet is proposed, which is a simple and unified encoder-decoder based FCN. 
FreeNet directly inputs the entire HSI without requiring any dimension reduction and outputs the classification map.
Therefore, FreeNet is a fully end-to-end trainable HSI classifier that does not require dimension reduction or any post-processing technology.
FreeNet avoids the redundant computation on the overlapping areas between patches, which significantly boosts its inference speed.
To maximize the exploitation of the global spatial context and details, a spectral attention module and lateral connection based SSF are proposed.
The spectral attention module models the interdependencies of the feature maps guided by the global spatial context to effectively boost the performance of FreeNet.
The lateral connection based SSF progressively refines the semantic features with the global spatial detail of the features from the shallow layers.
Meanwhile, lateral connection based SSF follows the residual learning approach to fuse the features by pointwise addition, which can alleviate the gradient vanishing problem, thereby, significantly improving the performance of FreeNet.

In the future, we will further explore memory-efficient HSI classifiers, which will be important for the real-time classification of HSIs from satellite and airborne platforms.
We hope that the proposed method will serve as a strong baseline and aid future research in HSI classification.

\section*{Acknowledgements}
The authors would like to thank the Editor, Associate Editor, and anonymous reviewers for their helpful comments and suggestions that improved this article.
% References should be produced using the bibtex program from suitable
% BiBTeX files (here: strings, refs, manuals). The IEEEbib.bst bibliography
% style file from IEEE produces unsorted bibliography list.
% -------------------------------------------------------------------------
\bibliographystyle{IEEEtran}
\bibliography{FPGA_arxiv}

\ifCLASSOPTIONcaptionsoff
  \newpage
\fi

\end{document}